\pdfoutput=1
\documentclass[11pt]{article}

\usepackage[dvipsnames]{xcolor}
\usepackage[]{emnlp2021}

\usepackage{times}
\usepackage{latexsym}

\usepackage[T1]{fontenc}
\usepackage[utf8]{inputenc}
\usepackage[scaled=0.8]{beramono}

\usepackage{microtype}

\usepackage{array}
\usepackage{subcaption}
\usepackage{graphicx}
\usepackage{amsfonts}
\usepackage{multirow}
\usepackage[linesnumbered,ruled,vlined]{algorithm2e}
\usepackage{amsmath}
\usepackage{amsthm}
\usepackage{amssymb}
\usepackage{bbm}
\usepackage{booktabs}
\usepackage[russian,greek,english]{babel}
\usepackage[normalem]{ulem}
\usepackage{tikz}
\usepackage{tikz-dependency}
\usetikzlibrary{shapes.geometric}
\usetikzlibrary{patterns}
\pgfdeclarepatternformonly{south west lines}{\pgfqpoint{-0pt}{-0pt}}{\pgfqpoint{3pt}{3pt}}{\pgfqpoint{3pt}{3pt}}{
        \pgfsetlinewidth{0.4pt}
        \pgfpathmoveto{\pgfqpoint{0pt}{0pt}}
        \pgfpathlineto{\pgfqpoint{3pt}{3pt}}
        \pgfpathmoveto{\pgfqpoint{2.8pt}{-.2pt}}
        \pgfpathlineto{\pgfqpoint{3.2pt}{.2pt}}
        \pgfpathmoveto{\pgfqpoint{-.2pt}{2.8pt}}
        \pgfpathlineto{\pgfqpoint{.2pt}{3.2pt}}
        \pgfusepath{stroke}}

\usepackage{pgfplots}
\pgfplotsset{compat=1.7}
\usetikzlibrary{backgrounds}
\usetikzlibrary{matrix,calc}

\tikzset{fit to page/.style={fit only width,fit only height},
         fit only width/.style={
                    trim left={($(current bounding box.center)-(0.5*\columnwidth,0)$)},
                    trim right={($(current bounding box.center)+(0.5*\columnwidth,0)$)},
          },
          fit only height/.style={
                     execute at end picture={%
                     \useasboundingbox let 
                           \p1=(current bounding box.north east),
                           \p2=(current bounding box.south west) in
                           \pgfextra{\pgfresetboundingbox}
                           (\x2,-0.5*\textheight) rectangle (\x1-(0,0.5*\textheight);
                    }
          }
}
\pgfplotsset{general plot/.style={
            ybar,
            every axis plot post/.style={/pgf/number format/fixed},
            bar width=.4cm,
            width=3cm,
            height=2cm,
            ymajorgrids=true,
            yminorgrids=false,
            xtick style={draw=none},
            xticklabels={,,},
            every y tick label/.append style={font=\small},
            tick pos=left,
            axis line style={draw=none},
            axis x line*=top,
            axis y line*=left,
            nodes near coords,
            every node near coord/.append style={font=\small,color=black},
            ymax=0,
            ytick={0},
            ylabel near ticks,
            enlarge x limits=1.0,
          }
}
\pgfplotsset{diachronic plot/.style={
            every axis plot post/.style={/pgf/number format/fixed},
            width=5cm,
            height=3cm,
            ymajorgrids=false,
            yminorgrids=false,
            xtick={'14,'15,'16,'17,'18,'19},
            xticklabels={14,15,16,17,18,19},
            every x tick label/.append style={font=\tiny},
            every y tick label/.append style={font=\tiny},
            ymax=1,ymin=0.7,
            tick pos=left,
            axis y line*=left,
            axis x line*=bottom,
            symbolic x coords={'14,'15,'16,'17,'18,'19},
            nodes near coords,
            every node near coord/.append style={font=\tiny,color=black},
            enlarge x limits=0.1,
            title style={yshift=-.1cm,font=\tiny},
    }
}

\pgfplotsset{wmt plot/.style={
            every axis plot post/.style={/pgf/number format/.cd,precision=2},
            width=5cm,
            height=3cm,
            ymajorgrids=false,
            yminorgrids=false,
            xtick={'14,'15,'16,'17,'18,'19},
            xticklabels={14,15,16,17,18,19},
            every x tick label/.append style={font=\tiny},
            every y tick label/.append style={font=\tiny},
            ymax=1,ymin=0,
            tick pos=left,
            axis y line*=left,
            axis x line*=bottom,
            symbolic x coords={'14,'15,'16,'17,'18,'19},
            every node near coord/.append style={font=\tiny,color=black,xshift=0pt,yshift=-0pt,anchor=north,},
            enlarge x limits=0.1,
            title style={yshift=-.5cm,font=\small},
    }
}

\newcommand{\tikplot}[2]{
\begin{tikzpicture}[trim left=0cm,trim right=0cm]
    \begin{axis}[diachronic plot,
    title={#2},]
    \addplot [mark=none] table[x=wmt,y=#1]{\diachronicdata};
    \addplot [only marks,mark=square*,mark options={fill=white,draw=white},mark size=1pt,] table[x=wmt,y=#1]{\diachronicdata};
    \addplot [only marks,mark=o,mark options={fill=blue,draw=blue},mark size=.3pt,] table[x=wmt,y=#1]{\diachronicdata};
    \end{axis}
\end{tikzpicture}
}
\newcommand{\tikplotshort}[2]{
\begin{tikzpicture}[trim left=0cm,trim right=0cm]
    \begin{axis}[diachronic plot,
    title={#2},height=2cm,ymin=0.9,ytick={0.9,1}]
    \addplot [mark=none] table[x=wmt,y=#1]{\diachronicdata};
    \addplot [only marks,mark=square*,mark options={fill=white,draw=white},mark size=1pt,] table[x=wmt,y=#1]{\diachronicdata};
    \addplot [only marks,mark=o,mark options={fill=blue,draw=blue},mark size=.3pt,] table[x=wmt,y=#1]{\diachronicdata};
    \end{axis}
\end{tikzpicture}
}
\newcommand{\tikplotmiddle}[2]{
\begin{tikzpicture}[trim left=0cm,trim right=0cm]
    \begin{axis}[diachronic plot,
    title={#2},height=2.5cm,ymin=0.8,ytick={0.9,1}]
    \addplot [mark=none] table[x=wmt,y=#1]{\diachronicdata};
    \addplot [only marks,mark=square*,mark options={fill=white,draw=white},mark size=1pt,] table[x=wmt,y=#1]{\diachronicdata};
    \addplot [only marks,mark=o,mark options={fill=blue,draw=blue},mark size=.3pt,] table[x=wmt,y=#1]{\diachronicdata};
    \end{axis}
\end{tikzpicture}
}

\newcommand{\wmtplot}[4]{
\begin{tikzpicture}[trim left=0cm,trim right=0cm,
    mycirctwo/.style={circle,solid,fill=black,draw=black,inner sep=0pt,minimum size=4pt},
    every pin/.style = {pin distance=2mm, 
                font=\scriptsize, inner sep=1pt},
    PNtwo/.style args = {#1/#2}{% Pin Node
    mycirctwo, pin=#1:#2}
    ]
    \begin{axis}[wmt plot, title={#3},height=4cm,ymin=#4,ytick={0.9,1}]
    \addplot [only marks,mark=o,mark options={fill=red!30,draw=red!50},mark size=1.5pt,] table[x=Year,y=Score]{#1};
    \addplot [only marks,mark=*,mark options={fill=black,draw=black},mark size=1.5pt,] table[x=year,y=ref]{#2};
    \addplot [mark=square*,mark options={fill=black,draw=black},mark size=2pt,nodes near coords,] table[x=year,y=avg]{#2};
\end{axis}
\end{tikzpicture}
}

\newcommand{\gn}[1]{\textcolor{cyan}{\bf\small [#1 --GN]}}
\newcommand{\ap}[1]{\textcolor{blue}{\bf\small [#1 --AP]}}

\newcommand{\drm}[1]{\textcolor{olive}{\bf\small [#1 --DRM]}}

\newcommand{\mname}{\textsc{l'ambre}}

\newcommand\blfootnote[1]{%
  \begingroup
  \renewcommand\thefootnote{}\footnote{#1}%
  \addtocounter{footnote}{-1}%
  \endgroup
}

\title{Evaluating the Morphosyntactic Well-formedness of Generated Texts}

\author{Adithya Pratapa,\textsuperscript{1*} Antonios Anastasopoulos,\textsuperscript{2*} Shruti Rijhwani,\textsuperscript{1} Aditi Chaudhary,\textsuperscript{1} \\ {\bf David R. Mortensen,\textsuperscript{1} Graham Neubig,\textsuperscript{1} Yulia Tsvetkov\textsuperscript{3}} \\
\textsuperscript{1}Language Technologies Institute, Carnegie Mellon University \\
\textsuperscript{2}Department of Computer Science, George Mason University \\
\textsuperscript{3}Paul G.~Allen School of Computer Science \& Engineering, University of Washington \\
\texttt{\{vpratapa,srijhwan,aschaudh,dmortens,gneubig\}@cs.cmu.edu} \\
\texttt{antonis@gmu.edu}, \texttt{yuliats@cs.washington.edu}
}

\begin{document}
\maketitle
\begin{abstract}
Text generation systems are ubiquitous in natural language processing applications. However, evaluation of these systems remains a challenge, especially in multilingual settings. In this paper, we propose \mname~-- a metric to evaluate the morphosyntactic well-formedness of text using its dependency parse and morphosyntactic rules of the language. We present a way to automatically extract various rules governing morphosyntax directly from dependency treebanks. To tackle the noisy outputs from text generation systems, we propose a simple methodology to train robust parsers. We show the effectiveness of our metric on the task of machine translation through a diachronic study of systems translating into morphologically-rich languages.\blfootnote{*Equal contribution}\footnote{Code and data are available at \url{https://github.com/adithya7/lambre}.}
% \blfootnote{$\dagger$ work done at Carnegie Mellon University.}
% \footnote{Code, data, and models will be released upon publication.}
% \sr{should we also mention the GEI analysis here?}
\end{abstract}

\section{Introduction}

A variety of natural language processing (NLP) applications such as machine translation (MT), summarization, and dialogue require natural language generation (NLG). Each of these applications has a different objective and therefore task-specific evaluation metrics are commonly used. For instance, reference-based measures such as BLEU~\cite{papineni-etal-2002-bleu}, METEOR~\cite{banerjee-lavie-2005-meteor} and chrF~\cite{popovic-2015-chrf} are used to evaluate MT, ROUGE~\cite{lin-2004-rouge} is a metric widely used in summarization, and various task-based metrics are used in dialogue \cite{liang-etal-2020-beyond}.

Regardless of the downstream application, an important aspect of evaluating language generation systems is measuring the fluency of the generated text. In this paper, we propose a metric that can be used to evaluate the \emph{grammatical well-formedness} of text produced by NLG systems.\footnote{While grammatical well-formedness is often necessary for fluent text, it is not sufficient \cite{sakaguchi-etal-2016-reassessing}.} Our metric is referenceless and is based on the grammatical rules of the language, thereby enabling fine-grained identification and analysis of which grammatical phenomena the NLG system is struggling with.

Although several referenceless metrics for evaluating NLG models exist, most use features of both the input and output, limiting their applicability to specific tasks like MT or spoken dialogue~\cite{specia2010machine,duvsek2017referenceless}.
%Some focus solely on evaluating fluency of the output~\cite{napoles-etal-2016-theres,novikova-etal-2017-need}.
With the exception of the grammaticality-based metric of ~\citet[GBM;][]{napoles-etal-2016-theres}, these metrics are derived from simple linguistic features like misspellings, language model scores or parser scores, and are not indicative of specific grammatical knowledge.

\begin{figure}[t]
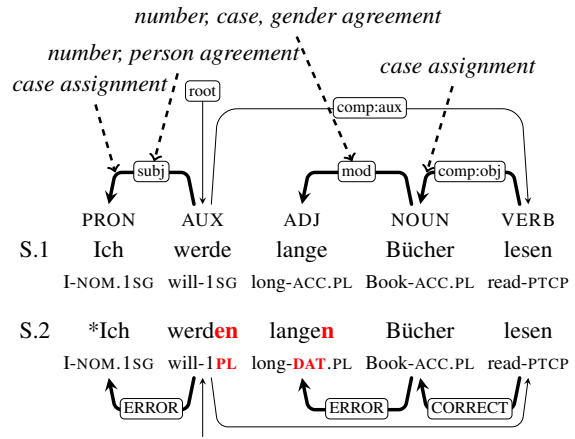

    \centering
    \resizebox{0.48\textwidth}{!}{
    \begin{dependency}[theme = default]
       \begin{deptext}[column sep=1pt, row sep=.2ex]
         \& \textsc{pron} \& \textsc{aux} \& \textsc{adj} \& \textsc{noun} \& \textsc{verb} \\
         S.1 \& Ich \& werde \& lange \& B\"{u}cher \& lesen \\
         \& \small I-\textsc{nom.1sg} \& \small will-\textsc{1sg} \& \small long-\textsc{acc.pl} \& \small Book-\textsc{acc.pl} \& \small read-\textsc{ptcp} \\[.3cm]
         S.2 \& *Ich \& werd\textcolor{red}{\textbf{en}} \& lange\textcolor{red}{\textbf{n}} \& B\"{u}cher \& lesen \\
         \& \small I-\textsc{nom.1sg} \& \small will-\textsc{1\textcolor{red}{\textbf{pl}}} \& \small long-\textsc{\textcolor{red}{\textbf{dat}}.pl} \& \small Book-\textsc{acc.pl} \& \small read-\textsc{ptcp} \\
       \end{deptext}
        
        \deproot{3}{root}
        \depedge[edge style={ultra thick}]{3}{2}{subj}
        \storelabelnode\subjedge
        \storesecondcorner\subjdepdedge
        \depedge{3}{6}{comp:aux}
        \depedge[edge style={ultra thick}]{6}{5}{comp:obj}
        \storesecondcorner\objedge
        \depedge[edge style={ultra thick}]{5}{4}{mod}
        \storelabelnode\modedge
        % \wordgroup{5}{2}{3}{subj}
        % \wordgroup{5}{4}{5}{mod}
        \deproot[hide label, edge below, edge unit distance=2ex]{3}{root}
        \depedge[edge style={ultra thick}, edge below]{3}{2}{ERROR}
        % \depedge[edge style={ultra thick}, edge below, edge height=1.2cm]{3}{2}{CORRECT}
        \depedge[hide label, edge below, edge height=0.8cm]{3}{6}{comp:aux}
        \depedge[edge style={ultra thick}, edge below]{6}{5}{CORRECT}
        \depedge[edge style={ultra thick}, edge below]{5}{4}{ERROR}
        % \depedge[edge style={ultra thick}, edge below, edge height=1.2cm]{5}{4}{CORRECT}

        \node (rule1) [above right of = \subjedge, xshift=-0.2cm, yshift=1.2cm] {\textit{number, person agreement \ \ \ }};
        \draw [->, dashed, very thick] (rule1) -- (\subjedge);
        \node (rule2) [above left of = \modedge, yshift = 1.8cm, xshift=-0.4cm] {\textit{number, case, gender agreement}};
        \draw [->, dashed, very thick] (rule2) -- (\modedge);
        \node (rule3) [above right of = \objedge, xshift=-0.2cm, yshift=1.0cm] {\textit{case assignment}};
        \draw [->, dashed, very thick] (rule3) -- (\objedge);
        \node (rule4) [above left of = \subjdepdedge, xshift=0.3cm, yshift=0.7cm] {\textit{case assignment}};
        \draw [->, dashed, very thick] (rule4) -- (\subjdepdedge);
    \end{dependency}}
    % \vspace{-.8cm}
    \caption{Identifying \textbf{\textcolor{red}{grammatical errors}} in text using dependency parses and morpho-syntactic rules. Ungrammatical sentence S.2 fails to satisfy subject-verb agreement between \textsc{pron} and \textsc{aux} as well as case agreement between \textsc{adj} and \textsc{noun}. However, it satisfies case assignment rules with the subject in \textsc{nom} case and the object in \textsc{acc} case respectively.}
    \label{fig:overview}
    % \vspace{-1em}
\end{figure}

% \gn{Where are the morpho-syntactic rules in the figure? What do green and red arrows mean? Also, green and red may not be the best color combination to use if the arrow colors are semantically meaningful, because this is the hardest color combination for colorblind people to see.}

% This technique cannot be used on the outputs of any black-box NLG system,

In contrast, there has recently been a burgeoning of evaluation techniques based on grammatical acceptability judgments for both language models \cite{marvin-linzen-2018-targeted,warstadt2019neural,gauthier-etal-2020-syntaxgym} and MT systems \cite{sennrich-2017-grammatical,burlot-yvon-2017-evaluating,burlot-etal-2018-wmt18}.
However, these methods require an existing model to score two sentences that are carefully crafted to be similar, with one sentence being grammatical and the other not.
These techniques are usually tailored towards specific %language models or translation 
downstream systems. Additionally, they do not consider the interaction between multiple mistakes that may occur in the process of generating text (e.g., an incorrect word early in the sentence may trigger a grammatical error later in the sentence).
Most of these methods, with the exception of \citet{mueller-etal-2020-cross}, focus only on English or translation to/from English. %, limiting their applicability to other languages.
% specific NLG systems \sr{what does "specific NLG systems" mean?}

In this paper, we propose \mname{}, a metric that \emph{both} evaluates the grammatical well-formedness of text in a fine-grained fashion and can be applied to text from multiple languages.
We use widely available dependency parsers %, such as Stanza~\citep{qi-etal-2020-stanza},
to tag and parse target text, and then compute our metric by identifying language-specific morphosyntactic errors in text (a schematic overview is outlined in \autoref{fig:overview}).
Our measure can be used \textit{directly} on text generated from a black-box NLG system, and allows for \emph{decomposing} the system performance into individual grammar rules that identify specific areas to improve the model's grammaticality.
% \drm{``Fluency'' is not technically incorrect here, but consider changing?}

\mname{} relies on a grammatical description of the language, similar to those linguists and language educators have been producing for decades when they document a language or create teaching materials.
Specifically, we consider rules describing morphosyntax, including agreement, case assignment, and verb form selection.
Following \citet{chaudhary-etal-2020-automatic}, we describe a procedure to automatically extract these rules from existing dependency treebanks (\S\ref{sec:grammatical_description}) with high precision.\footnote{While such sets of grammar rules could be manually compiled (for example, by linguists),
%the creators of these dependency treebanks, 
it would require additional centralized effort from a large group of annotators.}

When evaluating NLG outputs, adherence to these rules can be assessed through dependency parses (\autoref{fig:overview}).
However, off-the-shelf dependency parsers are trained on grammatically sound text and are not well-suited for parsing ungrammatical (or noisy) text~\cite{hashemi-hwa-2016-evaluation} such as that generated by NLG systems. We propose a method to train more robust dependency parsers and morphological feature taggers by synthesizing morphosyntactic errors in existing treebanks (\S\ref{sec:robust_parsing}). Our robust parsers improve by up to 2\% over off-the-shelf models on synthetically noised treebanks.
% \sr{where does this 2\% number come from? which table/figure? no need to mention it here, but I couldn't find it in Section 4}

Finally, we field test \mname{} on two NLP tasks: grammatical error identification (\S\ref{sec:gec_case_study}) and machine translation (\S\ref{sec:mt_case_study}). Our metric is highly correlated with human judgments on MT outputs. We also showcase how the interpretability of our approach can be used to gain additional insights through a diachronic study of MT systems from the Conference on Machine Translation (WMT) shared tasks. The success of our measure depends heavily on the quality of dependency parses: we discuss potential limitations of our approach based on the grammar error identification task.

\iffalse
\drm{At some point, we have to address a chicken-and-egg problem: to have a dependency parser, you need a treebank and to build a treebank, you need a certain amount of explicit grammatical knowledge (of things like agreement, since they can be diagnostic of dependency relations). We need to make an argument about how our methodology is better than just writing those rules down.} \ap{mentioned this in the rule extraction section} \gn{I still agree that ``writing rules is hard so we use a dependency parser'' seems a bit contradictory. Try to at least mention this a little in the intro.} \ap{added.}
\drm{I think it's much better in its current state.}
\fi

\section{\mname{}: Linguistically Aware Morphosyntax-Based Rule Evaluation}
\label{sec:metric}

In this section, we present \mname, a metric to gauge the morphosyntactic well-formedness of generated natural language sentences. 
Our metric assumes a \emph{machine-readable} grammatical description, which we define as a series of language-specific rules $G_l = \{r_1, r_2, \ldots, r_n\}$. We also assume that dependency parses of every grammatical sentence adhere to these rules.\footnote{Different syntactic formalisms could be applicable, but we work with the (modified) Universal Dependencies formalism~\cite{nivre-etal-2020-universal} due to its simplicity, widespread familiarity and its use in a variety of multilingual resources.}

Given a text, we compute a score by verifying the satisfiability of all applicable morphosyntactic rules from the grammatical description. Similar to standard metrics for evaluating NLG, our scoring framework allows for computing scores at both segment-level and corpus-level granularities.

\smallskip
\noindent
\textbf{Segment level:}\quad Computing \mname{} first requires segmentation, tokenization, tagging, and parsing of the corpus.\footnote{We discuss in \S\ref{sec:robust_parsing} how to properly achieve this over potentially malformed sentences.}
Given the tagged dependency tree for a segment of text and a set of rules in the language, we identify all rules that are applicable to the segment. We then compute the percentage of times that each such rule is satisfied within the segment, based on the parser/tagger annotations. The final score is a weighted average of the scores of individual rules.\footnote{We assume equal weights among rules, although it would be trivial to extend the metric to use a weighted average.} Our score lies between [0,1], where 1 and 0 represent that rules are perfectly satisfied or not satisfied at all respectively.
% with 0 and 1 indicating null and perfect satisfiability of applicable rules respectively.
% \sr{The explanation of how the metric is calculated is a bit unclear. The example is good, but it's not exactly clear what rules we are referring to that are correct or incorrect. Could we include an example of an incorrect rule within the text as well, so it's easier for the reader to understand the calculation with respect to the example?}
Consider the example sentence (S.2) from \autoref{fig:overview}. Of the five agreement rules, two rules, number agreement between \textsc{pron} (Ich) and \textsc{aux} (werde), and case agreement between \textsc{adj} (lange) and \textsc{noun} (B\"{u}cher) are not satisfied. Both relevant case assignment rules between, \textsc{pron} (Ich) and \textsc{aux} (werde), and \textsc{noun} (B\"{u}cher) and \textsc{verb} (lesen) are satisfied. Thus, the overall score is 0.71 (5/7). This example showcases how \mname{} is inherently interpretable: given a segment (S.2), we can immediately identify that it is grammatically sound with respect to case assignment, but contains two errors in agreement.

\smallskip
\noindent
\textbf{Corpus level:}\quad To compute \mname{} at corpus-level, we accumulate the satisfiability counts for each rule over the entire corpus and report the macro-average of the empirical satisfiability of each applicable rule. This is different from a simple average of segment-level scores and is a more reliable score as it allows the comparison of performance by rule over the entire corpus.%\\[.2cm]

% We now define our grammatical description to be used in the scoring framework.

\iffalse
As highlighted in this section, a formal grammatical description is an important component of our scoring framework. Next, we define the grammatical description and present a way to automatically compile it from dependency treebanks.
\fi

\section{Creating a Grammatical Description}
\label{sec:grammatical_description}

In linguistics, grammars of languages are typically presented in (series of) books, describing in detail the rules governing the language through free-form text and examples (see \citet{moravcsik1978agreement, corbett2006agreement} for grammatical agreement).\footnote{Many linguists also produce highly formal accounts of grammatical phenomena. However, many of these formalisms are difficult to implement computationally because they are equivalent (in the most egregious cases) to Turing machines.} 
%\drm{It is worth acknowledging that many linguists also produce highly formal accounts of grammatical phenomena. However, many of these formalisms are difficult to implement computationally because they are equivalent (in the most egregious cases) to Turing machines.} 
However, to be able to use such descriptions in our metric, we require them to be concise and \emph{machine-readable}.

We build upon~\citet{chaudhary-etal-2020-automatic} that constructed first-pass descriptions of grammatical agreement from syntactic structures of text, in particular, dependency parses.%
\footnote{We use the Surface-Syntactic Universal Dependencies (SUD) 2.5 \cite{gerdes-etal-2019-improving}. See \ref{ssec:appendix_ud_sud} for a comparison of UD and SUD.}
In general, rules based on a complete formalized grammar govern several aspects of language generation, including syntax, morphosyntax, morphology, morphophonology, and phonotactics. In this work, we focus on agreement, case assignment, and verb form choice.

\subsection{Agreement}

We define the \emph{agreement} rules as $r_{\textit{agree}}(x,y,d)\!\rightarrow\! f_x$=$f_y$. Such rules refer to two words with parts-of-speech $x$ (dependent) and $y$ (head/governer) connected through a dependency relation $d$. These two words must exhibit agreement on some morphological feature $f$. For instance, the noun \textit{B\"{u}cher} (`Book') and its modifying adjective \textit{lange} (`long') in the German example S.1 (\autoref{fig:overview}) agree in number, gender, and case. We denote this general agreement rule as $r_{\textit{agree}}(\mathtt{ADJ},\mathtt{NOUN},\mathtt{mod})\rightarrow \mathtt{Case},\mathtt{Gender},\mathtt{Number}$.

For each dependency relation $d$ between a dependent POS $x$ and head POS $y$, we compute the fraction of times the linked tokens agree on feature $f$ in the treebank. We consider $r_{\textit{agree}}(x,y,d)\rightarrow f$ as a potential agreement rule if the fraction is higher than 0.9. The resulting set still contains a long tail of less-frequent rules. These are unreliable and could just be because of treebank artifacts. Therefore, we incorporate additional pruning to only select the most frequent rules, covering a cumulative 80\% of all agreement instances in the treebank. This is a simplified formulation compared to \citet{chaudhary-etal-2020-automatic}, but as we show later, this frequency-based approach still results in a high-precision set of rules.

\subsection{Case Assignment and Verb Form Choice}

We define case assignment and verb form choice rules as $r_{\textit{as}}(x,y,d)\!\rightarrow\!f_x$=$F$. A word with POS $x$ at the tail of a dependency relation $d$ with head POS $y$ must exhibit a certain morphological feature (i.e., $f_x$ must have the value~$F$). Occasionally, a similar rule might be applicable for the head $y$. For instance, a pronoun that is the child of a \texttt{subj} relation (that is, it is the subject of a verb) in most Greek constructions must be in the nominative case, while a direct object (\texttt{obj}) should be in the accusative case. In this example, we can write the rules as $r_{\textit{as}}(\mathtt{PRON},\mathtt{VERB},\mathtt{subj})\!\rightarrow\!\mathtt{Case_{PRON}}\!=\!\mathtt{Nom}$ and $r_{\textit{as}}(\mathtt{PRON},\mathtt{VERB},\mathtt{obj})\!\rightarrow\!\mathtt{Case_{PRON}}\!=\!\mathtt{Acc}$. 

\iffalse
This class of rules are also often lexicalized, depending on the lexeme of either the head or the dependent. In the example S.1 of Figure~\ref{fig:overview}, the object phrase `lange B\"ucher' is inflected in the accusative case because of the verb `lesen'. Other constructions might require the object placed in genitive or dative, depending on the lexeme of the verb.\footnote{We note that there are argument structure alternations in verbs that affect the case of arguments without a necessary change in the lexeme.} However, we leave the extension to lexicalized rules as future work.
\fi

\begin{figure}[t]
    %\centering
\pgfplotstableread[row sep=\\,col sep=&]{
val & call & cobj  \\
Nom & 30.4 & 11.3 \\
Gen & 11.5 & 0.4 \\
Acc & 24.8 & 86.7 \\
Dat & 33.3 & 1.6\\
}\dedata
\def\mystrut{\vphantom{h}}
\begin{tikzpicture}[trim left=-0.5cm,trim right=0cm]
    \begin{axis}[
            ybar,
            every axis plot post/.style={/pgf/number format/fixed},
            bar width=.6cm,
            width=8.5cm,
            height=3.5cm,
            ymajorgrids=false,
            yminorgrids=false,
            %every axis legend/.code={\let\addlegendentry\relax},
            legend style={draw=none,fill=none,at={(1,1.1)},anchor=north east},
            xtick={Nom, Gen, Acc, Dat},
            symbolic x coords={Nom, Gen, Acc, Dat},
            every x tick label/.append style={font=\mystrut},
            every y tick label/.append style={font=\small\mystrut},
            tick pos=left,
            %hide y axis,
            axis y line*=left,
            axis x line*=bottom,
            nodes near coords,
            nodes near coords align={vertical},
            every node near coord/.append style={font=\small,color=black},
            %nodes near coords style={},
            %title={Case Distribution in German},
            %title style={yshift=0cm},
            ymin=0,ymax=100,
            ylabel shift={-1cm},
            ylabel near ticks,
            %ylabel={},
            %ylabel near ticks,
            %xlabel={}
            enlarge x limits=0.2,
        ]
        %\addplot [style={black,postaction={pattern=north east lines},fill=white,mark=none}] table[x=story,y=base]{\ainudata};
        \addplot [style={blue,mark=none},postaction={pattern=south west lines,pattern color=blue}] table[x=val,y=call]{\dedata};
        \addplot [style={Dandelion,fill=Dandelion,mark=none},] table[x=val,y=cobj]{\dedata};
        \legend{$G(f)$, $L(f)$};
        \node[draw=none] at (50,90){\small{$\text{KL}(G,L)\!=\!0.9122$}};
    \end{axis}
\end{tikzpicture}
    \caption{Argument structure rules: the global case distribution $G(f)$ of German NOUN is very different from its local $L^{\textit{depd}}(f)$ distribution in \textit{comp:obj} dependency with VERB, allowing us to the identify case assignmnt rule $r_{\textit{as}}(\mathtt{NOUN},\!\mathtt{VERB},\!\mathtt{comp\!:\!obj})$\!$\rightarrow$\!$\mathtt{Case_{NOUN}}\!=\!\mathtt{Acc;Nom}$, i.e the Case can be either Acc or Nom.}
    \label{fig:asr_example}
    % \vspace{-1em}
\end{figure}
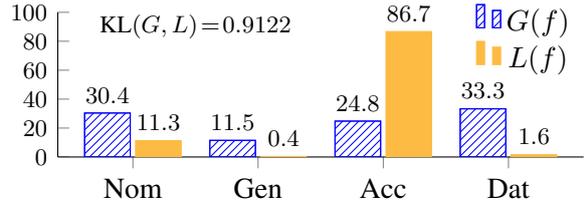

% \an{The notation doesn't match, but we'll probably upadte the text} \an{The rule should probably only involve Acc and not Nom?} \ap{The frac. instances with Nom is > 10\%, can't ignore?} \an{It seems very counter-intuitive. Nom in general has 30\% --> under this rule, it has 11\%. I would expect that anything that gets (significantly) *higher* that its G(f) val would be the interesting rule one.}

% To automatically identify rules pertaining to case assignment and verb form choice, we compute three different feature-value distributions from the treebanks. 
Our hypothesis is that certain syntactic constructions require specific morphological feature selection from one of their constituents (e.g., pronoun subjects \textit{need} to be in nominative case, but pronoun objects only allow for genitive or accusative case in Greek).\footnote{This class of rules are also often lexicalized, depending on the lexeme of either the head or the dependent. In the example S.1 of \autoref{fig:overview}, the object phrase \textit{lange B\"ucher} (`long Book') is inflected in the accusative case because of the verb \textit{lesen} (`read'). Other constructions might require the object declined in genitive or dative, depending on the verb lexeme.}
This implies that the ``local'' distribution that a specific construction requires will be different from a ``global'' distribution of morphological feature values computed over the whole treebank. \autoref{fig:asr_example} presents an example for German-GSD.

% \drm{It is my preference that we following the linguistic convention for citing forms and glosses: Forms that you are talking about in text (e.g. \textit{B\"ucher}) are given in italics. Glosses (translations) are given in single quotes. When you give a form in a foreign language, it is also polite to provide a gloss.}

\iffalse
For instance, in the German GSD treebank, we notice a stark difference in the global distribution of noun case and its local distribution within a \textit{comp:obj} relation with a verb (see Figure \ref{fig:asr_example}), helping us identify this morphosyntactic rule in German grammar. In our German example from Figure \ref{fig:overview}, the noun B\"{u}cher satisfies this rule.
\fi

We can automatically discover these rules by finding such cases of distortion. First, we obtain a global distribution ($G(f_x) = p(f_x)$) that captures the empirical distribution of the values of a morphological feature $f$ on POS $x$ over the whole treebank. Second, we measure two other distributions, local to a relation $d$, for the dependent ($L^{\textit{depd}}(f_x \mid d) = p(f_x \mid \langle x,*,d \rangle)$) and head positions ($L^{\textit{head}}(f_x \mid d) = p(f_x \mid \langle *,x,d \rangle)$)

\iffalse
\begin{align*}
    G(f_x) &= p(f_x) \\
    L^{\textit{depd}}(f_x \mid d) &= p(f_x \mid \langle x,*,d \rangle) \\
    L^{\textit{head}}(f_x \mid d) &= p(f_x \mid \langle *,x,d \rangle)
\end{align*}    
\fi

To identify these morphosyntactic rules with high precision, we measure the KL divergence~\cite{kullback1951information} between global and local distributions and only keep the rules with KL divergence over a predefined threshold of 0.9. Similar to the case of agreement rules, we impose a frequency threshold on the count of dependency relation in the respective treebank. For all the agreement, case assignment and verb form choice rules, we use the largest SUD treebank for the language.

\subsection{Human Evaluation}

Though our grammatical description incorporates agreement, case assignment and verb-form selection, which are highly indicative of the fluency of natural language text, it is by no means exhaustive. However, these rules are relatively easy to extract from dependency parses with high precision.
To measure the quality of our extracted rule sets, we perform a human evaluation task with three linguists.\footnote{Disclaimer: One annotator is also an author on this work.} Similar to \citet{chaudhary-etal-2020-automatic}, for each rule, we present three choices, ``\textit{almost always true}'', ``\textit{sometimes true}'' and ``\textit{need not be true}'', along with 10 positive and negative examples from the original treebank.\footnote{Due to large number of Russian rules, we present only a subset to the linguists.} In \autoref{tab:human_eval_results}, we show the results for Greek, Italian and Russian.

\begin{table}[t]
    \centering
    \begin{tabular}{lc@{ \ }c@{ \ }c}
    \toprule
    \textbf{Rules} & \textbf{Greek} & \textbf{Russian} & \textbf{Italian} \\
    \midrule
    $r_{\textit{agree}}$ & 11:0:0 & 17:3:0 & - \\
    $r_{\textit{as}}$ & 9:3:0 & 6:9:3 & 10:1:0 \\
    \bottomrule
    \end{tabular}
    \caption{Results on human evaluation of our automatically extracted rules. Numbers denotes \# rules labeled as (\textit{always}):(\textit{sometimes}):(\textit{need not})}
    \label{tab:human_eval_results}
    % \vspace{-1em}
\end{table}

\iffalse
\begin{table}[t]
    \centering
    \begin{tabular}{l|c|@{ \ }c|@{ \ }c}
    \toprule
    \textbf{Language} & \textbf{always} & \textbf{sometimes} & \textbf{not rules} \\
    \midrule
    \multicolumn{4}{c}{Agreement rules} \\    
    %\midrule
    Greek & 11 & 0 & 0 \\
    Russian & 17 & 3 & 0 \\
    \midrule
    \multicolumn{4}{c}{Case assignment and VerbForm choice rules} \\    
    %\midrule
    Greek & 9 & 3 & 0 \\
    Italian & 10 & 1 & 0 \\
    Russian & 6 & 9 & 3 \\
    \bottomrule
    \end{tabular}
    \caption{Results on human evaluation of our automatically extracted rules.}
    \label{tab:human_eval_results}
    \vspace{-1em}
\end{table}
\fi

Our rules are in general quite precise across the three languages, with most rules marked as ``almost always true'' by linguists. However, we found interesting special cases in Russian, where the annotator stated that dependency relations are ``overloaded'' to capture several phenomena (explaining the ``sometimes'' annotations). The SUD schema merges \texttt{obj} and \texttt{ccomp} into a single \texttt{comp:obj} relation, thereby we notice instances where the rule $r_{\textit{as}}(\mathtt{PRON},\mathtt{VERB}, \mathtt{comp}\mathtt{:}\mathtt{obj})\rightarrow\mathtt{Case}_\mathtt{PRON}=\mathtt{Acc}$ (which pertains to direct objects) is incorrectly enforced on a \texttt{ccomp} relation. We also notice some issues with cross-clausal dependencies, e.g., the rule $r_{\textit{as}}(\mathtt{VERB},\mathtt{NOUN}, \mathtt{subj})\rightarrow\mathtt{VerbForm}_\mathtt{VERB}=\mathtt{Inf}$ is valid in the sentence, ``the \textit{goal} is to \textit{win}'' but not in ``the \textit{question} is why they \textit{came}''. 

It is important to note that these automatically extracted rule sets are approximate descriptions of morpho-syntactic behavior of the language.
However, \mname{} is flexible enough to utilize any additional rules, and arguably would be even more effective if combined with hand-curated descriptions created by linguists.
We leave this as an interesting direction for future work. In our code, we provide detailed instructions for adding new rules.
% \sr{you can attach code (anonymized) to the submission if that would help here}.

% Before using our rules in \mname{}, we first evaluate off-the-shelf parsers on ungrammatical text.

\iffalse
With high-quality rules at hand and before presenting applications of \mname{} on NLP tasks, we first tackle the issue of off-the-shelf parsers performing poorly on ungrammatical text~\cite{hashemi-hwa-2016-evaluation}. In the next section, we illustrate this issue in our use of parsers and then propose a method to easily train robust parsers.
\fi

% \drm{I like these signposting paragraphs but they do take up space. Perhaps they need to be reduced/eliminated?}

\section{Parsing Noisy Text}
\label{sec:robust_parsing}

Within our evaluation framework, we rely on parsers to generate the dependency trees of potentially malformed or noisy sentences from NLG systems. However, publicly available parsers are typically trained on clean and grammatical text from UD treebanks, and may not generalize to noisy inputs~\cite{daiber-van-der-goot-2016-denoised, sakaguchi-etal-2017-error, hashemi-hwa-2016-evaluation, hashemi2018jointly}. Therefore, it is necessary to ensure that parsers are robust to any morphology-related errors in the input text. Ideally, the tagger should accurately identify the morphological features of incorrect word forms, while the dependency parser remains robust to such noise. To this end, we present a simple framework for evaluating the robustness of pre-trained parsers to such noise, along with a method to train the robust parsers necessary for our application. 

\subsection{Adding Morphology-related Noise}
\label{ssec:noise}

\begin{figure}[t]
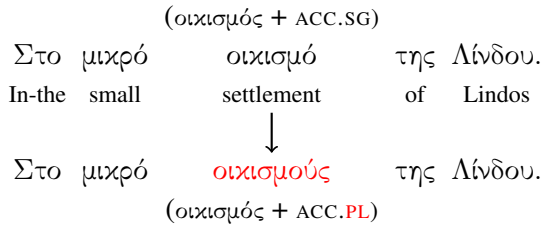

    \centering
    %\resizebox{0.5\textwidth}{!}{
    \begin{dependency}[theme = default]
        \begin{deptext}[column sep=1pt, row sep=.2ex]
            \& \& (\foreignlanguage{greek}{\small{οικισμός}} + \small{\textsc{acc.sg}}) \& \& \\
            \foreignlanguage{greek}{Στο} \& \foreignlanguage{greek}{μικρό} \& \foreignlanguage{greek}{οικισμό} \& \foreignlanguage{greek}{της} \& \foreignlanguage{greek}{Λίνδου.} \\
            \small{In-the} \& \small{small} \& \small{settlement} \& \small{of} \& \small{Lindos} \\ \\
            \foreignlanguage{greek}{Στο} \& \foreignlanguage{greek}{μικρό} \& \foreignlanguage{greek}{\textcolor{red}{οικισμούς}} \& \foreignlanguage{greek}{της} \& \foreignlanguage{greek}{Λίνδου.} \\
            \& \& (\foreignlanguage{greek}{\small{οικισμός}} + \small{\textsc{acc.\textcolor{red}{pl}}}) \& \& \\
        \end{deptext}
        
        % \node[text width=2.5cm, align=left, above left of = \wordref{1}{3}, xshift=-1cm, yshift=0.7cm, draw, rounded corners, fill=black!5] (ann) {\small{lemma:}\hskip 5pt \foreignlanguage{greek}{\small{οικισμός}}\\\small{feats:}\hskip 5pt \small{\textsc{acc.sg}}};
        % \draw [->, thick] (\wordref{1}{3}) -- (ann);
        
        % \node[text width=3cm, align=left, above right of = \wordref{1}{3}, xshift=1.1cm, yshift=0.7cm, draw, rounded corners, fill=black!5] (alt) {\small{\textsc{nom.sg}:}\hskip 2pt\small{\foreignlanguage{greek}{οικισμός}}
        % \small{\textsc{gen.sg}:}\hskip 4pt\small{\foreignlanguage{greek}{οικισμού}}
        % \small{\textsc{acc.pl}:}\hskip 6pt\small{\foreignlanguage{greek}{οικισμούς}}};
        % \draw [->, thick] (alt) -- (\wordref{1}{3});
        % \draw [->, thick] (ann) -- (alt);
        \draw[->, thick] (\wordref{3}{3}) -- (\wordref{5}{3});
        
    \end{dependency}
    %}
    % \vspace{-1em}
    \caption{Creating noisy input examples for parsers. In this Greek example, we modify the original word form \foreignlanguage{greek}{οικισμό} (Singular) to a plural inflection
    %from UniMorph dictionary, 
    \foreignlanguage{greek}{οικισμούς}.}
    \label{fig:noise_example}
    % \vspace{-1em}
\end{figure}
To simulate noisy input conditions for parsers, we add morphology-related errors into the standard UD treebanks using UniMorph dictionaries~\cite{mccarthy-etal-2020-unimorph}. UniMorph provides a schema for inflectional morphology by listing paradigms with relevant morphological features from an universal schema~\cite{sylak2016composition}. Given an input sentence, we search for alternate inflections for the constituent tokens, based on their lemmata.\footnote{For each token, we first map the morphological feature annotations in the original UD schema to the UniMorph schema~\cite{mccarthy-etal-2018-marrying}.}
%\footnote{\url{https://github.com/unimorph/ud-compatibility/blob/master/UD_UM/UD-UniMorph.tsv}}.
For simplicity, we only replace a single token in each sentence and for this token, we substitute with a form differing in exactly one morphological feature (e.g., Case, Number, etc.). 
For each sentence in the original treebank, we sample a maximum of one altered sentence. \autoref{fig:noise_example} illustrates the construction of a noisy (or altered) version of an example sentence from Greek-GDT treebank.\footnote{\citet{tan-etal-2020-morphin} follows similar methodology using English-only \texttt{LemmInflect} tool, but our approach is scalable to the large number of languages in UniMorph.}

% \footnote{Recent work~\cite{tan-etal-2020-morphin} follows a similar method to induce morphology related errors for English using the English-only \texttt{lemmainflect} tool. In comparison, our approach scales to a large number of languages (UniMorph is constantly expanding and currently supports more than 100 languages).}

In general, we were able to add noise to more than 80\% of the treebanks' sentences, but in a few cases we were constrained by the number of available paradigms in UniMorph (see \ref{ssec:appendix_robust_parsing} for more details). A potential solution could utilize a inflection model like the \texttt{unimorph\_inflect} package of \citet{anastasopoulos-neubig-2019-pushing}, but we leave this for future work.

% For example, the Turkish dictionary contains just 3.5k paradigms as compared to 28k in Russian, and we could only corrupt about 55\% of the Turkish sentences.
% \footnote{\url{https://stanfordnlp.github.io/stanza/available_models.html}}

For evaluation, we induce noise into the \textit{dev} portions of the treebanks and test the robustness of off-the-shelf taggers and parsers from Stanza \cite{qi-etal-2020-stanza} (indicative results on Czech, Greek, and Turkish are shown in \autoref{fig:robust_parsing_results}).
Along with the overall scores on the dev set, we also report the results only on the altered word forms (``Altered Forms''). Across the three languages, we notice a significant drop in tagger performance, with a more than 30\% drop in feature tagging accuracy of the altered word forms. The parsing accuracy is also affected, in some cases significantly. This reinforces observations in prior work and illustrates the need to build more robust parsers and taggers.

\input{images/parsing_results}

\subsection{Training Robust Parsers}

To adapt to the noisy input conditions in practical NLP settings like ours, our proposed solution is to re-train the parsers/taggers directly on noisy UD treebanks. With the procedure described above (\S\ref{ssec:noise}) we also add noise to the \textit{train} splits of the UD v2.5 treebanks and re-train the lemmatizer, tagger, and dependency parser from scratch.\footnote{We added errors into UD, and not SUD, as it allows for re-using the original Stanza hyperparameters, and also facilitates for application of robust parsers outside of \mname{}. Note that, conversion between UD and SUD can be done with minimal loss of information.} To retain the performance on clean inputs, we concatenate the original clean train splits with our noisy ones. 
We experimented with commonly used multilingual parsers like UDPipe~\cite{straka-strakova-2017-tokenizing}, UDify~\cite{kondratyuk-straka-2019-75}, and Stanza~\cite{qi-etal-2020-stanza}, settling on Stanza for its superior performance in preliminary experiments. 
We use the standard training procedure that yields state-of-the-art results on most UD languages with the default hyperparameters for each treebank. %\footnote{\url{https://stanfordnlp.github.io/stanza/training.html}} 
Given that we are inherently tokenizing the text to add morphology-related noise, we reuse the pre-trained tokenizers instead of retraining them on noisy data.

% \drm{``performance on [...] results'' sounds bad to me. Maybe just ``superior preliminary results''?}

% \footnote{While we report only LAS, we notice similar gains on UAS. LAS is more relevant to our application because of the need for labeled dependency links.}

\autoref{fig:robust_parsing_results} compares the performance of the original and our robust parsers on three treebanks. Overall, we notice significant improvements on both LAS (with similar gains on UAS) and UFeat accuracy on the altered treebank as well as the altered forms. 
%It is worth noting that, additionally, 
Importantly, our robust parsers retain the state-of-the-art performance on clean text.
In all the analyses reported henceforth (unless explicitly mentioned), we use our \textit{robust} Stanza parsers trained with the above-described procedure.

\section{Does \mname{} Capture Grammaticality?}
\label{sec:gec_case_study}

Before deploying \mname{} on automatically generated text, we need to ensure that our approach is indeed able to identify syntactic ill-formedness.
Grammar error correction (GEC) datasets are an ideal test bed. In its original formulation, the GEC task involves identifying and correcting errors relating to spelling, morphosyntax and word choice. For evaluating \mname{}, we only focus on \textit{grammar error identification} (GEI) and specifically on identification of morphosyntactic errors.
%  related to word form
% Because of this special focus, we don't present a comparison to other GEI techniques, but it is an interesting direction for future work.

% Note, \mname{} is not directly applicable for correcting errors it identifies.

We experiment with two morphologically rich languages, Russian and German. We use the Falko-MERLIN GEC corpus~\cite{boyd-2018-using} for German and the RULEC-GEC dataset~\cite{rozovskaya-roth-2019-grammar} for Russian. We focus on error types related to morphology (see \ref{ssec:appendix_gec}). %An example sentence from the German dataset, ``\textit{Vielleicht verdienen Ärzte \sout{ihre} ihr \sout{aktuelle} aktuelles Gehalt}'', illustrates a case assignment error.
% \footnote{We are limited to these languages because error-tagged GEC data are not available in any other languages, except for English.}
% Table~\ref{tab:gec_dataset_examples} presents an example with a case assignment error from the German dataset in the standard M2 annotation format.

% \input{images/gec_examples}

\paragraph{Evaluation:}
\label{ssec:gec_eval}

To evaluate the effectiveness of \mname{}, we run it on the \emph{training}\footnote{We use the train portion due to its large size, therefore gives a better estimate of our \mname{} performance. Note that, in this experiment, we do not aim to compare against state-of-the-art GEI tools.} splits of the German and Russian GEC datasets. 
GEC corpora typically annotate single words or phrases as errors (and provide a correction); in contrast, we only identify errors over a dependency link, which can then be mapped over to either the dependent or head token. This difference is not trivial: a subject-verb agreement error, for instance, could be fixed by modifying either the subject or the verb to agree with the other constituent.
To account for this discrepancy, we devise a schema to ensure the proper computation of precision and recall scores. First, we detect any errors at a given token by evaluating all the valid \mname{} rules between the curren token and its dependency neighbors (head, dependents). If there is a gold error at the current token, we consider it a true positive or false negative depending on whether or not we detect the error. For false positive cases, we divide the score between the current token and the neighbor via the erroneous dependency link (see \autoref{algo:gec_eval_method} in \ref{ssec:appendix_gec}).

\autoref{tab:gec_results} presents the results using both agreement ($r_\textit{agree}$) and argument structure rules (case assignment and verb form choice, $r_\textit{as}$).

% \footnote{Due to paucity of space, we skip the exact computation details here.}

\iffalse
\begin{table}[t]
    \centering
    \begin{tabular}{@{}llcc@{}}
    \toprule
    \multirow{2}{*}{\textbf{Language}}& \multirow{2}{*}{\textbf{Parser}} & \multicolumn{2}{c}{$r_\textit{agree} \cup r_\textit{as}$} \\
    %\cmidrule{2-3}
    && Precision & Recall \\
    \midrule
    %\multicolumn{3}{l}{\textbf{German GEC}} \\
    %\midrule
    \multirow{2}{*}{German} & Original & 35.5 & 24.9 \\
     & Robust & 31.4 & 30.4 \\
     & Robust++ & 38.3 & 29.9 \\
    \midrule
    %\multicolumn{3}{l}{\textbf{Russian GEC}} \\
    %\midrule
    \multirow{2}{*}{Russian} & Original & 18.5 & 20.9 \\
    & Robust & 18.5 & 25.7 \\
    \bottomrule
    \end{tabular}
    \caption{Precision and Recall of morphosyntactic errors on train splits of German and Russian GEC.} %We present results using both off-the-shelf Stanza models as well as the proposed robust parsers.}
    \label{tab:gec_results}
\end{table}
\fi

\begin{table}[t]
    \centering
    \small
    \begin{tabular}{@{}l@{ \ }l@{}c@{}c@{}c@{ \ }c@{ \ }c@{ \ }c@{}}
    \toprule
    \multirow{2}{*}{\textbf{Lang.}}& \multirow{2}{*}{\textbf{Parser}} & \multicolumn{2}{l}{$r_\textit{agree} \cup r_\textit{as}$} & \multicolumn{2}{l}{$r_\textit{agree}$} & \multicolumn{2}{l}{$r_\textit{as}$} \\
    %\cmidrule{2-3}
    && P & R & P & R & P & R \\
    \midrule
    %\multicolumn{3}{l}{\textbf{German GEC}} \\
    %\midrule
    % \multirow{3}{*}{German} & Original & 35.5 & 24.9 & 37.5 & 23.4 & \textbf{28.7} & \textbf{2.5} \\
    %  & Robust & 31.4 & \textbf{30.4}  & 36.3 & \textbf{30.4} & 15.1 & \textbf{2.8}\\
    %  & Robust++ \ & \textbf{38.3} & 29.9 & \textbf{46.5} & 28.3 & 15.1 & \textbf{2.8}\\
    \multirow{3}{*}{German} & Original & 32.6 & 29.6 & 34.2 & 28.9 & \textbf{14.9} & \textbf{1.0} \\
     & Robust & 33.6 & \textbf{34.5}  & 35.2 & \textbf{33.8} & 12.2 & 0.8 \\
     & Robust++ \ & \textbf{40.0} & 34.1 & \textbf{42.5} & 33.4 & 12.2 & 0.8\\
    \midrule
    %\multicolumn{3}{l}{\textbf{Russian GEC}} \\
    %\midrule
    \multirow{2}{*}{Russian} & Original & \textbf{18.5} & 20.9 & \textbf{22.6} & 18.7 & 9.9 & 3.7 \\
    & Robust & \textbf{18.5} & \textbf{25.7} & 22.1 & \textbf{20.6} & \textbf{14.6} & \textbf{8.1} \\
    \bottomrule
    \end{tabular}
    \caption{Precision and Recall of morphosyntactic errors on train splits of German and Russian GEC. Robust++ indicates results after additional manual post-correction of rules.} %We present results using both off-the-shelf Stanza models as well as the proposed robust parsers.}
    \label{tab:gec_results}
    % \vspace{-1em}
\end{table}

\paragraph{Analysis:}
\label{ssec:gec_discussion}

In both languages, we find agreement rules to be of higher quality than case and verb form assignment ones. This phenomenon is more pronounced in German where many case assignment rules are lexeme-dependent, as discussed in~\S\ref{sec:grammatical_description}.

Importantly, our proposed \textbf{robust parsers lead to clear gains in error identification recall}, compared to the pre-trained ones (``Original'' vs. ``Robust'' in \autoref{tab:gec_results}).
Given the complexity of the errors present in text from non-native learners and the well-known incompleteness of GEC corpora in listing all possible corrections~\cite{napoles-etal-2016-theres}, combined with the prevalence of typos and the dataset's domain difference compared to the parser's training data, our error identification module performs quite well.

To understand where \mname{} fails, we manually inspected a sample of false positives. First, we notice that tokens with typos are often erroneously tagged and parsed. Our augmentation is only equipped to handle (correctly spelled) morphological variants. %Incorporating character-level noise or 
Additionally applying a spell checker might be beneficial in future work.

\iffalse
We manually inspected a sample of false positives from our model's annotations, to understand where \mname{} fails. First, we notice that tokens with typos are often erroneously tagged and parsed. Since our training data augmentation technique only focused on (correctly spelled) morphological variants, we did not anticipate our parsers to be robust on typos. We expect that adding character-level noise in the training data would be beneficial, but we leave this extension for future work.
\fi

Second, we find that German interrogative sentences and sentences with more rare word order (e.g., object-verb-subject) are often incorrectly parsed, leading to misidentifications by \mname. In the training portion of the German HDT treebank, 76\% of the instances present the subject before the verb and the object appears after the verb in 62\% of the sentences. Questions and subordinate clauses that follow the reverse pattern (OVS order) make a significant portion of the false positives.

Last, we find that morphological taggers exhibit very poor handling of syncretism (i.e., forms that have several possible analyses), often producing the most common analysis regardless of context. 
For example, nominative-accusative  syncretism is well documented in modern German feminine nouns~\cite{krifka2003case}. German auxiliary verbs like \textit{werden} (`will') that share the same form for 1st and 3rd person plurals, are almost always tagged with the 3rd person. As a result, our method mistakenly identifies correct pronoun-auxiliary verb subject dependency constructions as violations of the rule $r_{agree}(\mathtt{PRON},\mathtt{AUX},\mathtt{subj})\!\rightarrow\!\mathtt{Person}$, as the \textsc{pron} and \textsc{aux} are tagged with disagreeing person features (1st and 3rd respectively). 
By manually correcting for this issue over our German rules (by specifically discounting such cases) we improve \mname's precision by almost~7 percentage points (``Robust++'' in \autoref{tab:gec_results}).

\paragraph{Comparison with Other Metrics:}

We also compare \mname{} to other metrics that capture fluency and/or grammatical well-formedness, namely perplexity as computed by large language models and the grammaticality-based metric (GBM) of \citet{napoles-etal-2016-theres}. To provide a fair comparison of \mname{}, perplexity and GBM, we reformulate the GEI task into an acceptability judgment task. Specifically, we check if the metrics' score the grammatical target sentence higher than the ungrammatical source sentence in the GEI test split for Russian and German. To compute the perplexity scores, we use transformer-based LMs \cite{ng-etal-2019-facebook}. GBM relies on the open-source Language Tool \cite{milkowski2010developing}, which is a widely-used rule-based proofreading software to detect sentence-level errors.\footnote{\url{https://languagetool.org/dev}} GBM measures error count rate as $1 - \frac{\#\text{errors}}{\#\text{tokens}}$. For additional details on the task setup, we refer the readers to \ref{ssec:appendix_gec_evaluation} in Appendix.
% We provide concrete details and exhaustive results on GEI test splits in Table~\ref{tab:gec_contrast} in Appendix~\ref{ssec:appendix_gec_evaluation}.

Our findings are two-fold. First, perplexity performs better than the other metrics. However, perplexity cannot provide any error diagnosis, so it by itself is not useful for providing feedback to a user. Second, \mname{} is better at capturing morpho-syntactic rules necessary for grammatical correctness (especially in Russian), while GBM is better at other fluency-related aspects. While both \mname{} and GBM are interpretable, \mname{}'s UD-based rule construction makes it easier to extend it to new languages.\footnote{In \mname{}, we reuse the expertise of UD annotators, but in GBM, expanding to a new language requires native speakers to craft new regexes.} For complete results, refer to \autoref{tab:gec_contrast} in Appendix~\ref{ssec:appendix_gec_evaluation}.
% \drm{``Rules'' is the wrong word, I think.} 

In our GEI analysis, we utilized the Russian and German GEC corpora for evaluating the quality of \mname{}. In future work, it would be interesting to expand the analysis to datasets from other languages, Czech \cite{naplava-straka-2019-grammatical} and Ukrainian \cite{syvokon2021uagec}.

\section{Evaluating NLG: A Machine Translation Case Study}
\label{sec:mt_case_study}

% We illustrate an application of \mname{} to evaluate a wide variety of MT systems. While, in principle, \mname{} can be applied for evaluating fluency of generic NLG outputs
Grammaticality measures, including \mname{}, can be useful across NLG tasks. Here, we chose MT due to the wide-spread availability of (human-evaluated) system outputs in many languages. %We leave the application to other NLG tasks to future work.

In addition to BLEU, chrF and t-BLEU\footnote{t-BLEU~\cite{Ataman2020A} measures BLEU on outputs tagged using a morphological analyzer.} are commonly used to evaluate translation into morphologically-rich languages~\cite{goldwater-mcclosky-2005-improving, toutanova-etal-2008-applying, chahuneau-etal-2013-translating,sennrich-etal-2016-neural}.
% Along with the standard BLEU, metrics like chrF3~\cite{popovic-2015-chrf} and t-BLEU\footnote{t-BLEU~\cite{Ataman2020A} measures BLEU on outputs tagged using a morphological analyzer.} are commonly used to evaluate translation into morphologically-rich languages
Evaluating the well-formedness of MT outputs has previously been studied \cite{popovic-etal-2006-morpho}. Recent WMT shared tasks included special test suites to inspect linguistic properties of systems~\cite{sennrich-2017-grammatical,burlot-yvon-2017-evaluating,burlot-etal-2018-wmt18}, which construct an evaluation set of contrastive source sentence pairs (typically English). While such contrastive pairs are very valuable, they only \textit{implicitly} evaluate well-formedness and require access to underlying MT models to score the contrastive sentences. In contrast, \mname{} \emph{explicitly} measures well-formedness, without requiring access to trained MT models.

% Evaluating the syntactic well-formedness of MT system outputs has been of growing interest over the last decade~\cite{popovic-etal-2006-morpho}. Since 2018, the WMT has included test suites to inspect linguistic properties of systems submitted to the News Translation Shared Tasks~\cite{ma-etal-2018-results,ma-etal-2019-results}, such as LingEval97~\cite{sennrich-2017-grammatical}, MorphEval~\cite{burlot-yvon-2017-evaluating} and the extended MorphEval~\cite{burlot-etal-2018-wmt18}. All these approaches construct an evaluation set of contrastive sentence pairs in the source language (typically English) and evaluate translation systems based on the generated sentence pair in the target language. These evaluation sets also include a few pairs to test agreement.

% Although manually curated source-side contrastive pairs are very valuable, they only \textit{implicitly} evaluate the fluency of the generated outputs by comparing to references. In contrast, \mname{} directly measures syntactic well-formedness without requiring access to the trained MT models. Other related works identify grammatical errors in translation outputs by comparing the dependency trees of reference and hypothesis segments~\cite{duma-etal-2013-new,tezcan-etal-2016-detecting}. 
% \mname{} instead can evaluate any NLG outputs in the desired target language as it does not rely on references.

 %and does not require direct access to the trained MT models.

% \footnote{\url{http://www.statmt.org/wmt19/metrics-task.html}}
% \footnote{\url{http://www.statmt.org/wmt19/translation-task.html}} 
For evaluating MT systems, we use the data from the Metrics Shared Task in WMT 2018 and 2019~\cite{ma-etal-2018-results,ma-etal-2019-results}. This corpus includes outputs from all participating systems on the test sets from the News Translation Shared Task~\cite{bojar-etal-2018-findings,barrault-etal-2019-findings}.
Our study focuses on systems that translate from English to morphologically-rich target languages: Czech, Estonian, Finnish, German, Russian, and Turkish. We used all relevant languages from the WMT shared task except for Lithuanian and Kazakh, which lack reasonable quality parsers.

%In the following section, we first present a correlation analysis of our proposed metric with the human judgments (or direct assessments) for systems from WMT18 and WMT19. We then present a case-study of machine translation into morphologically-rich languages through a diachronic analysis of MT systems from WMT14 to WMT19.

% \input{images/wmt_correlations}
\begin{table}[t]
    \centering
    \small
    \newcommand\dottedcircle{\raisebox{-.1em}{\tikz \draw [line cap=round, line width=0.2ex, dash pattern=on 0pt off 0.5ex] (0,0) circle [radius=0.79ex];}}
    \resizebox{0.48\textwidth}{!}{
    \begin{tabular}{@{}l|c|c|c|c|c|c@{}}
    \toprule
    en$\rightarrow$\dottedcircle{}%$\dagger$ 
    & cs & de & et & fi & ru & tr \\
    \midrule
    \multicolumn{7}{l}{\textbf{WMT'18}} \\
    %\midrule
    all & 0.84 & -0.06 & 0.68 & 0.86 & 0.86 & 0.58 \\
    $r_\textit{agree}$ & 0.91 & 0.07 & 0.83 & 0.96 & 0.71 & 0.64 \\
    $r_\textit{as}$ & 0.78 & -0.10 & 0.62 & 0.77 & 0.89 & -0.31 \\
    \midrule
    \multicolumn{7}{l}{\textbf{WMT'19}} \\
    %\midrule
    all & 0.80 & 0.16 & - & 0.85 & 0.57 & - \\
    $r_\textit{agree}$ & 0.89 & 0.14 & - & 0.87 & 0.70 & - \\
    $r_\textit{as}$ & 0.70 & 0.13 & - & 0.82 & 0.45 & - \\
    \bottomrule
    \end{tabular}}
    \caption{With a few exceptions, our grammar-based metrics correlate well with human evaluations of WMT18 and WMT19 systems (Pearson's $r$ against Z-scores). Results using robust Stanza parsers.}
    \label{tab:wmt_correlations}
    % \vspace{-1em}
\end{table}

\iffalse
\begin{table}[t]
    \centering
    \newcommand\dottedcircle{\raisebox{-.1em}{\tikz \draw [line cap=round, line width=0.2ex, dash pattern=on 0pt off 0.5ex] (0,0) circle [radius=0.79ex];}}
    \resizebox{0.5\textwidth}{!}{
    \begin{tabular}{@{}l|c|c|c|c|c|c@{}}
    \toprule
    en$\rightarrow$\dottedcircle{}$\dagger$ & cs & de & et & fi & ru & tr \\
    \midrule
    WMT'18 & 0.84 & -0.08 & 0.68 & 0.86 & 0.86 & 0.59 \\
    WMT'19 & 0.80 & 0.16 & - & 0.85 & 0.57 & - \\
    \bottomrule
    \end{tabular}}
    \caption{With a few exceptions, our grammar-based metrics correlate well with human evaluations of WMT18 and WMT19 systems. (Pearson's $r$ against Z-scores). $\dagger$:we remove outlier systems following \citet{mathur-etal-2020-tangled}. Results using robust Stanza parsers.}
    \label{tab:wmt_correlations}
    \vspace{-1em}
\end{table}
\fi

\paragraph{Correlation Analysis}
\label{ssec:mt_correlation_analysis}
The MT system outputs are accompanied with human judgment scores, both at the segment and system level. 
%First, we measure the correlation of our corpus-level metric (\S\ref{sec:metric}) with the system-level human judgments.
In contrast to the reference-free nature of human judgments, our scorer is both reference-free \textit{and} source-free.

Following the standard WMT procedure for evaluating MT metrics, we measure the Pearson's $r$ correlations between \mname{} and human z-scores for systems from WMT18 and WMT19.
We follow~\citet{mathur-etal-2020-tangled} to remove outlier systems, since they tend to significantly boost the correlation scores, making the correlations unreliable, especially for the best performing systems~\cite{ma-etal-2019-results}.
%Though Pearson's $r$ correlation has been the standard for metrics evaluation and system comparisons, recent work~\cite{ma-etal-2019-results} has indicated that its behavior is unstable, especially for the best translation systems. In fact, as~\citet{mathur-etal-2020-tangled} highlight, outlier systems tend to boost the correlation score significantly. Therefore, we follow the method proposed by~\citet{mathur-etal-2020-tangled} to remove outlier systems for each language.
\autoref{tab:wmt_correlations} presents the correlation results for WMT18 and WMT19.\footnote{See \ref{ssec:appendix_wmt_correlation_study} for the corresponding scatter plots.}

We generally observe moderate to high correlation with human judgments using both sets of rules across all languages, apart from German (WMT18,19). This confirms that grammatically sound output is an important factor in human evaluation of NLG outputs. The correlation is lower with case assignment and verb form choice rules, with notable negative correlations for German, and Turkish (WMT18). In the case of German, a significant number of case assignment rules are dependent on the lexeme (as noted in \S\ref{sec:grammatical_description}) and we expect future work on lexicalized rules to partially address this drawback. In Turkish, the low parser quality plays a significant role and highlights the need for further work on parsing morphologically-rich languages~\cite{tsarfaty-etal-2020-spmrl}. %Beyond this, we attribute the few low correlations to very similar fluency \mname{} scores across the systems. This leads to low correlation with human judgments which incorporate both fluency and adequacy (with respect to the source). 
%The otherwise high correlations, despite the fact that human judgments, unlike \mname, incorporate both fluency \textit{and} adequacy with respect to the source, denote that grammatically-sound output is an important factor in human evaluations.
Last, we note that human judgments, unlike \mname, incorporate both well-formedness \textit{and} adequacy (with respect to the source). Therefore, we recommend using \mname{} in tandem with standard MT metrics to obtain a good indication of overall performance, both during model training and evaluation.

%For completeness, 
We additionally perform a correlation analysis of \mname{} with perplexity, BLEU and chrF on the WMT system outputs (\ref{ssec:appendix_wmt_correlation_study} in Appendix). As expected, we see a strong negative correlation with perplexity (low perplexity and high \mname{}). For BLEU and chrF, the results are quite similar to the correlations with human z-scores.

\input{images/wmt}

\begin{figure}
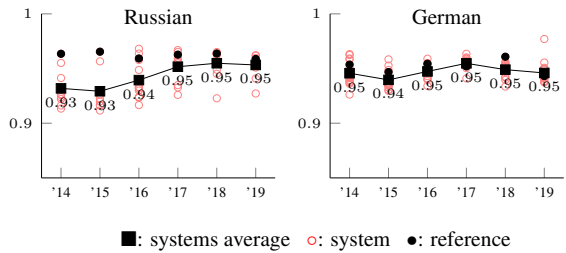

\centering
\resizebox{0.48\textwidth}{!}{
\begin{tabular}{p{3.8cm}p{3.8cm}}
\wmtplot{\rudata}{\ruavg}{Russian}{.85} & \wmtplot{\dedata}{\deavg}{German}{.85} \\[.3em]
\multicolumn{2}{c}{\small $\blacksquare$: systems average \ \ \textcolor{red}{$\circ$}: system \ \ $\bullet$: reference}\\
%\wmtplot{\csdata}{\csavg}{Czech}{.9}
%\wmtplot{\fidata}{\fiavg}{Finnish}{.85}
%\wmtplot{\trdata}{\travg}{Turkish}{.95}\\
\end{tabular}}
% \vspace{-.5em}
    \caption{A diachronic study of grammatical well-formedness of WMT English$\rightarrow$X systems' outputs. The systems in general are becoming more fluent. In the last two years the best systems produce as well-formed outputs as the reference translations.} %\ap{should we report median or mean score?}}
    \label{fig:diachronic_results}
    % \vspace{-1em}
\end{figure}

\paragraph{Diachronic Analysis}
\label{ssec:mt_diachronic_analysis}

We present an additional application of \mname{} through a diachronic study of translation systems submitted to the WMT news translation tasks. We run our scorer on system outputs from WMT14~\cite{bojar-etal-2014-findings} to WMT19~\cite{barrault-etal-2019-findings} for translation models from English to German and Russian.\footnote{Similar analysis on Czech, Finnish and Turkish in \ref{ssec:appendix_diachronic_wmt}.} \autoref{fig:diachronic_results} shows the scores of all systems and highlights the average trend of system scores. We also present the scores on the reference translations for comparison. We observe that systems have gotten more fluent over the years, often as good as the reference translations in the most recent shared tasks.\footnote{Such comparison of well-formedness scores is reasonable, to an extent, even though test sets differ year-to-year, as we measure the grammatical acceptability but not adequacy.}
%Also note that unlike the previous setting of correlation computation, we present the scores for all the systems including the outliers.

\pgfplotstableread[row sep=\\,col sep=&]{
wmt & agr-conj-NOUN-NOUN:Case & agr-det-DET-NOUN:Gender & agr-flat@name-PROPN-PROPN:Number & agr-mod-ADJ-NOUN:Case & agr-mod-VERB-NOUN:Case & agr-mod-VERB-NOUN:Gender & agr-mod-VERB-NOUN:Number & agr-subj-NOUN-VERB:Gender & agr-subj-NOUN-VERB:Number & agr-subj-PRON-VERB:Person & agr-subj@pass-NOUN-VERB:Number & args-comp:obj-NOUN-VERB:depd:Case & args-comp:obj@x-VERB-ADJ:depd:VerbForm & args-comp:obj@x-VERB-VERB:depd:VerbForm & args-subj-NOUN-AUX:depd:Case & args-subj-NOUN-VERB:depd:Case & args-subj@pass-NOUN-AUX:depd:Case & args-udep-PRON-VERB:depd:Case & args-udep-VERB-NOUN:depd:VerbForm\\ 
'14 & 0.7789 & 0.9894 & 0.9974 & 0.9258 & 0.8244 & 0.9050 & 0.8525 & 0.9202 & 0.8719 & 0.9930 & 0.8592 & 0.9522 & 1.0000 & 1.0000 & 0.9521 & 0.9389 & 0.8909 & 0.8684 & 1.0000 \\ 
'15 & 0.7760 & 0.9861 & 1.0000 & 0.9123 & 0.8364 & 0.9041 & 0.8628 & 0.9085 & 0.8584 & 0.9884 & 0.8110 & 0.9522 & 1.0000 & 1.0000 & 0.9201 & 0.9275 & 0.9006 & 0.9028 & 1.0000 \\ 
'16 & 0.8088 & 0.9906 & 1.0000 & 0.9405 & 0.8392 & 0.9199 & 0.8779 & 0.9252 & 0.8700 & 0.9902 & 0.8521 & 0.9684 & 1.0000 & 1.0000 & 0.9451 & 0.9232 & 0.8889 & 0.9221 & 1.0000 \\ 
'17 & 0.8848 & 0.9900 & 1.0000 & 0.9714 & 0.9345 & 0.9300 & 0.9500 & 0.9670 & 0.9275 & 0.9972 & 0.9241 & 0.9804 & 1.0000 & 1.0000 & 0.9556 & 0.9364 & 0.8900 & 0.9259 & 1.0000 \\ 
'18 & 0.8962 & 0.9930 & 1.0000 & 0.9745 & 0.9332 & 0.9336 & 0.9470 & 0.9800 & 0.9378 & 0.9955 & 0.9367 & 0.9840 & 1.0000 & 1.0000 & 0.9619 & 0.9398 & 0.8846 & 0.9272 & 1.0000 \\ 
'19 & 0.8636 & 0.9928 & 1.0000 & 0.9758 & 0.9268 & 0.9424 & 0.9423 & 0.9718 & 0.9362 & 0.9972 & 0.9315 & 0.9873 & 1.0000 & 1.0000 & 0.9749 & 0.9386 & 0.9032 & 0.8734 & 1.0000 \\ 
}\diachronicdata

\begin{figure}
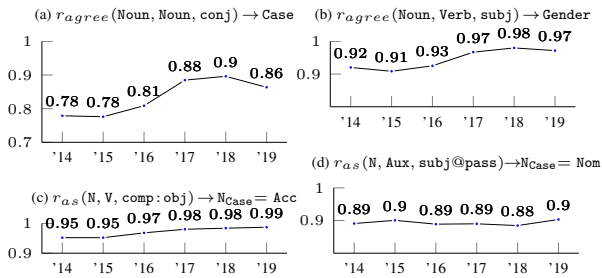

\centering
\resizebox{0.48\textwidth}{!}{
\begin{tabular}{m{3.6cm}m{3.6cm}}
\begin{tabular}{c}
\tikplot{agr-conj-NOUN-NOUN:Case}{(a) $r_{agree}(\mathtt{Noun,Noun,conj})\!\rightarrow\!\mathtt{Case}$}\\
\tikplotshort{args-comp:obj-NOUN-VERB:depd:Case}{(c) $r_{as}(\mathtt{N,V,comp\!:\!obj})\!\rightarrow\!\mathtt{N}_{\mathtt{Case}}\mathtt{=Acc}$}\\
\end{tabular}
&
\begin{tabular}{c}
\tikplotmiddle{agr-subj-NOUN-VERB:Gender}{(b) $r_{agree}(\mathtt{Noun,Verb,subj})\!\rightarrow\!\mathtt{Gender}$}
\\
\tikplotmiddle{args-subj@pass-NOUN-AUX:depd:Case}{(d) $r_{as}(\mathtt{N,Aux,subj@pass})\!\!\rightarrow\!\!\mathtt{N}_{\mathtt{Case}}\mathtt{=Nom}$}\\
%\tikplotshort{}{$r_{agree}(\mathtt{Adj,Noun,mod})\!\rightarrow\!\mathtt{Case}$}\\
\end{tabular}
\end{tabular}}
%  \vspace{-.5em}
\caption{Diachronic analysis of select agreement ($r_{agree}$) and argument structure ($r_{as}$) rules in Russian. We report \textit{median} well-formedness score per year. WMT systems have consistently improved on their well-formedness, but some phenomena are still challenging, such as handling agreement across conjuncted nouns (a) or %proper 
casing in passive constructions (d).}
\label{fig:diachronic_results_rules}
% \vspace{-1em}
\end{figure}
\mname{} also allows for fine-grained analysis of NLG systems by identifying specific grammatical issues. We illustrate this through a diachronic comparison of WMT systems for English$\rightarrow$Russian on a subset of \mname{}'s morphosyntactic rules (\autoref{fig:diachronic_results_rules}), presenting the median score per rule and year. % For simplicity, we use the median fluency score of systems as the representative score for a given WMT year.
Such fine-grained analysis reveals interesting trends. For example, while systems have been performing well on some rules over the years (\autoref{fig:diachronic_results_rules} (c)), there are rules that improved only in recent years (\autoref{fig:diachronic_results_rules} (a)). We also identify rules for constructions that remain challenging even for the best systems from WMT19 (\autoref{fig:diachronic_results_rules}~(d)). %Similar insights can be easily derived for other target languages.

\section{Conclusion and Future Work}
\label{sec:discussion}

% \sr{not sure why only the conclusion was in past tense, I edited it but feel free to switch back if needed}
In this paper, we introduce \mname{}, a framework to evaluate grammatical acceptability of text by verifying morphosyntactic rules over dependency parse trees. We present a method to automatically extract such rules for many languages along with a method to train robust parsing models which facilitate better verification of these rules on natural language text. We demonstrate the practical application of \mname{} on the popular generation task of machine translation, focusing on translation into morphologically-rich languages. 
Directions for future work include (1) incorporating additional morphosyntactic rules (e.g., word order), automatically extracted or hand-crafted ones such as those in \citet{mueller-etal-2020-cross} and (2) building more robust parsers and morphological taggers that are aware of the dependency structure of the sentence.

%For future work, we identify two main areas of interest. Firstly, the grammatical rule sets can be expanded to include other morphosyntactic rules (e.g. word order), automatically extracted or hand-crafted ones a la~\citet{mueller-etal-2020-cross}.
%This would allow for even more in-depth analysis of NLG systems. 
%Secondly, as illustrated in our analysis of grammatical error identification task, it is important to build robust morphological feature taggers that are aware of the dependency structure of the sentence, which has the potential to aid in tagging and parsing text in morphologically rich languages and further improve \mname{}'s robustness.% Such awareness will lead to better identification of agreement and argument structure within the text.

\section*{Acknowledgments}

The authors would like to thank Maria Ryskina for help with human evaluation of extracted rules, and Alla Rozovskaya for sharing the Russian GEC corpus with us.
This work was supported in part by the
National Science Foundation under grants 1761548, 2007960, and 2125201. Shruti Rijhwani was supported by a Bloomberg Data Science Ph.D. Fellowship.
This material is partially based on research sponsored by the Air Force Research Laboratory under agreement number FA8750-19-2-0200. The U.S. Government is authorized to reproduce and distribute reprints for Governmental purposes notwithstanding any copyright notation thereon. The views and conclusions contained herein are those of the authors and should not be interpreted as necessarily representing the official policies or endorsements, either expressed or implied, of the Air Force Research Laboratory or the U.S. Government.
%%% Ack Maria, Alla Rozovskaya

\bibliography{anthology,custom}
\bibliographystyle{acl_natbib}

\clearpage
\newpage

\appendix

\section{Appendix}
\label{sec:appendix}

\subsection{Comparison of UD and SUD}
\label{ssec:appendix_ud_sud}

\begin{figure}[ht]
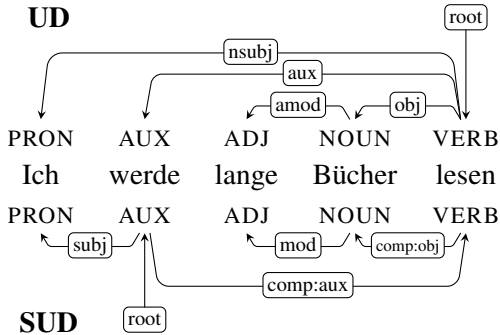

    \centering
    %\resizebox{0.5\textwidth}{!}{
    \begin{dependency}[theme = default, edge horizontal padding=2pt]
       \begin{deptext}[column sep=8pt, row sep=.2ex]
         \textsc{pron} \& \textsc{aux} \& \textsc{adj} \& \textsc{noun} \& \textsc{verb} \\
         Ich \& werde \& lange \& B\"{u}cher \& lesen \\
         \textsc{pron} \& \textsc{aux} \& \textsc{adj} \& \textsc{noun} \& \textsc{verb} \\
       \end{deptext}
        
        \deproot[edge height=45pt]{5}{root}
        \depedge[edge height=25pt]{5}{1}{nsubj}
        \storesecondcorner\nsubj
        \depedge[edge height=17pt]{5}{2}{aux}
        \depedge[edge height=5pt]{5}{4}{obj}
        \depedge[edge height=5pt]{4}{3}{amod}
        %\deproot[edge below, edge height=5pt]{2}{root}
        \depedge[edge below, edge height=5pt]{2}{1}{subj}
        \storesecondcorner\subj
        \storelabelnode\subjedge
        \depedge[edge below, edge height=20pt]{2}{5}{comp:aux}
        \depedge[edge below, edge height=5pt]{5}{4}{\small {comp:obj}}
        \depedge[edge below, edge height=5pt]{4}{3}{mod}
        \deproot[edge below, edge unit distance=10pt]{2}{root}
        
        \node (ud) [above of = \nsubj, yshift=-15pt] {\textbf{UD}};
        %\draw [->, dashed, very thick] (ud) -- (\nsubj);
        
        \node (sud) [below of = \subj] {\textbf{SUD}};
        %\draw [->, dashed, very thick] (sud) -- (\subj);
        
        % \node (rules) [below left of = \subjedge, yshift=-10pt] {\textit{case,person agreement}};
        %\draw [->, dashed, very thick] (rules) -- (\subjedge);
        
    \end{dependency}
    %}
    \caption{The SUD tree (below) for the sentence ``Ich werde lange B\"ucher lesen'' links the auxiliary verb ``werde'' with its subject ``Ich'' capturing an agreement rule not present in the UD tree (above).}
    \label{fig:ud_sud_example}
    \vspace{-1em}
\end{figure}

A comparison of the UD and SUD trees for the German sentence from \autoref{fig:overview} is presented in \autoref{fig:ud_sud_example}. Unlike the UD parse, the SUD parse directly links the \texttt{PRON} and \texttt{AUX}, allowing for an easy inference of relevant morphosyntactic rules.

% reference: https://en.wikibooks.org/wiki/LaTeX/Linguistics#Two-dimensional_Dependency_Trees

\subsection{Robust parsing}
\label{ssec:appendix_robust_parsing}

We proposed a methodology to utilize UniMorph dictionaries to add morphology-related noise into UD treebanks. Sometimes, the amount of noise we can add is limited by the number of available paradigms in UniMorph. For example, the Turkish dictionary contains just 3.5k paradigms as compared to 28k in Russian, and we could only corrupt about 55\% of the Turkish sentences.

\subsection{GEC datasets}
\label{ssec:appendix_gec}

In our evaluation on GEC, we only select morphology-related errors in German and Russian GEC datasets. Specifically, we use all errors of the type $\langle$\textsc{POS}$\rangle$:form from German Falko-MERLIN GEC corpus. In the Russian RULEC-GEC dataset, we select errors of types Case (Noun, Adj), Number (Noun, Verb, Adj), Gender (Noun, Adj), Person (Verb), Aspect (Verb), Voice (Verb), Tense (Verb), Other (Noun, Verb, Adj) and word form.

The methodology for computing the precision and recall in our GEC evaluation (from \autoref{tab:gec_results}) is presented in \autoref{algo:gec_eval_method}.

\begin{algorithm}[t]
\SetAlgoLined
\KwResult{P = $\frac{tp}{tp+fp}$, R = $\frac{tp}{tp+fn}$}
\caption{GEC using extracted morphosyntactic rules. \\
$H(t), G(t)$: estimated and gold error in token $t$ \\
$h^{(t)}, D^{(t)}$: head and dependents of token $t$ \\
$f_r(h^{(t)}, t)$: rule $r$ is satisfied in the dependency link between $t$ and $h^{(t)}$ \label{algo:gec_eval_method}}
$tp, fp, fn = 0, 0, 0$~\;
\For{sent in doc}{
    \For{$t$ in sent}{
        $E(t) = \Big\{t^{*}: \neg f_r(t^{*}, t), \forall t^{*} \in h^{(t)} \cup D^{(t)} \Big\}$~\;
        $H(t) = len(E(t)) > 0$~\;
        \If{$H(t)$}{
            \eIf{$G(t)$}{
                $tp$ += 1~\;
            }{
                \For{$t^*$ in $E(t)$}{
                    \If{$\neg G(t^*)$}{
                        $fp$ += 0.5~\;
                    }
                }
            }
        }
        \If{$G(t) \land \neg H(t)$}{
            $fn$ += 1~\;
        }
    }
}
\end{algorithm}

% \textsc{Noun}:Case,Number,Gender ; \textsc{Verb}:Number,Person ; \textsc{Verb}:Aspect,Voice,Tense ; \textsc{Adj}:Case;Gender;Number, \textsc{Noun/Verb/Adj}:Other and word form.

\subsection{GEC evaluation}
\label{ssec:appendix_gec_evaluation}

\paragraph{Comparison with Other Metrics}
We additionally  present a comparison of \mname{} to other metrics that capture fluency and/or grammatical well-formedness. One metric is perplexity, computed by large language models (LM). Specifically, we use transformer-based LMs~\cite{ng-etal-2019-facebook}. %\footnote{\url{https://github.com/pytorch/fairseq/blob/master/examples/language_model/README.md}} 
Second, we use a grammaticality-based metric (GBM) \cite{napoles-etal-2016-theres} that relies on the open-source Language Tool \cite{milkowski2010developing}. Language Tool is a widely-used rule-based proofreading software used to detect sentence-level errors. GBM measures error count rate as $1 - \frac{\#\text{errors}}{\#\text{tokens}}$. 

To provide a fair comparison of these three methods, we first reframe the GEI task into an acceptability judgment task. Given a source sentence from GEC corpus, we prepare four variants using the annotations provided with the corpus, 1. source sentence itself (no corrections made), 2. morph-corrected sentence (only morphology related corrections are made), 3. rest-corrected sentence (only non-morphology related corrections are made), and 4. target sentence (all corrections made).

To evaluate the effectiveness of the three metrics, we make 5 contrastive comparisons as shown in \autoref{tab:gec_contrast}. For instance, in the comparison (src, tgt), $\forall$ src $\neq$ tgt, we check if \mname{}(src) $<$ \mname{}(tgt), GBM(src) $<$ GBM(tgt) and PPL(src) $>$ PPL(tgt).\footnote{These strict inequalities allow us to capture limitations of rule-based methods. An error might be undetectable if the corresponding rule is absent in the method's rule set.} \autoref{tab:gec_contrast} presents the accuracy results across the 5 contrastive pairs on the \emph{test} splits of German and Russian GEC corpora. Overall, perplexity performs much better than the other metrics across all pairs. However, unlike GBM and \mname{}, perplexity doesn't provide error diagnosis, with no feedback on incorrect grammatical rules. Between \mname{} and GBM, former is competitive or better at two pairs, (src, morph-corrected) and (rest-corrected, tgt), whereas the latter does better at two other pairs, (src, rest-corrected), (morph-corrected, tgt). These results indicate that the proposed metric, \mname{}, is good at capturing morpho-syntactic rules necessary for grammatical correctness (especially in Russian), and the more complex GBM does better at other fluency related rules. Additionally, we observed clear improvements by using our proposed robust parsers (\S\ref{sec:robust_parsing}) over the original stanza parsers.

\begin{table}[t]
    \centering
    \begin{tabular}{@{}lc@{ \ }c@{ \ }c@{}}
    \toprule
    Contrast & \mname{} & GBM & PPL \\
    \midrule
    \multicolumn{4}{l}{\textbf{German}} \\
    %\midrule
    (src, tgt) & 0.30 (0.28) & \underline{0.63} & \textbf{0.95} \\
    (src, morph-corrected) & 0.31 (0.29) &\underline{0.32} & \textbf{0.70} \\
    (src, rest-corrected) & 0.21 (0.19) & \underline{0.61} & \textbf{0.92} \\
    (morph-corrected, tgt) & 0.20 (0.18) & \underline{0.62} & \textbf{0.96} \\
    (rest-corrected, tgt) & 0.41 (0.39) & \underline{0.47} & \textbf{0.97} \\
    \midrule
    \multicolumn{4}{l}{\textbf{Russian}} \\
    %\midrule
    (src, tgt) & 0.21 (0.20) & \underline{0.40} & \textbf{0.94} \\
    (src, morph-corrected) & \underline{0.24} (0.22) & 0.14 & \textbf{0.74} \\
    (src, rest-corrected) & 0.12 (0.12) & \underline{0.43} & \textbf{0.88} \\
    (morph-corrected, tgt) & 0.18 (0.16) & \underline{0.46} & \textbf{0.95} \\
    (rest-corrected, tgt) & \underline{0.35} (0.33) & 0.18 & \textbf{0.94} \\
    \bottomrule
    \end{tabular}
    \caption{Accuracy results for \mname{}, grammaticality-based metric (GBM), and perplexity (PPL) on various contrastive acceptability judgments on German and Russian GEC (\textit{test} splits). Best score is in \textbf{bold}, and the second best score is \underline{underlined}. Numbers in parentheses are obtained by using original Stanza parsers instead of the proposed robust parsers.}
    \label{tab:gec_contrast}
\end{table}

In our re-implementation of the GBM, we follow the prior work \cite{napoles-etal-2016-theres} and utilize Language Tool (LT) for error detection. We use LT for two languages, German (de-DE: Germany) and Russian (ru-RU). The source sentences in both the GEC corpora are pre-tokenized, therefore, we skip whitespace-based rules while using LT. For Russian, we remove whitespace-based rules corresponding to comma, punctuation and hypen. For German, we remove whitespace-based rules corresponding to quotation mark, exclamation mark, unit spaces, comma and parentheses. In the German GEC test split, the total counts of each contrastive pairs (x, y) with sent(x) $\neq$ sent(y), (src, tgt): 1791, (src, morph-corrected): 1169, (src, rest-corrected): 1646, (morph-corrected, tgt): 1582, and (rest-corrected, tgt): 831. In the Russian GEC split, the total counts of each contrastive pairs (x, y) with sent(x) $\neq$ sent(y), (src, tgt): 2381, (src, morph-corrected): 1405, (src, rest-corrected): 2005, (morph-corrected, tgt): 1913, and (rest-corrected, tgt): 1148.

\subsection{WMT Correlation Studies}
\label{ssec:appendix_wmt_correlation_study}

\begin{table}[ht]
    \centering
    \newcommand\dottedcircle{\raisebox{-.1em}{\tikz \draw [line cap=round, line width=0.2ex, dash pattern=on 0pt off 0.5ex] (0,0) circle [radius=0.79ex];}}
    \begin{tabular}{@{}r@{}c|c|c|c@{}}
    \toprule
    \small{English}$\rightarrow$\dottedcircle{} \ & \#sys$^\dagger$ & $r_\textit{agree}$ & $r_\textit{as}$ & $r_\textit{agree} \cup r_\textit{as}$ \\
    \midrule
    \multicolumn{5}{@{}l}{\textbf{WMT'18}}\\
    Czech & 5 & 0.91 & 0.78 & 0.84 \\
    % German & 12 & 0.22 & 0.51 & 0.45 \\
    German & 12 & 0.07 & -0.10 & -0.06 \\
    Estonian & 12 & 0.83 & 0.62 & 0.68 \\
    Finnish & 12 & 0.96 & 0.77 & 0.86 \\
    Russian & 7 & 0.71 & 0.89 & 0.86 \\
    Turkish & 7 & 0.64 & -0.31 & 0.58 \\
    \midrule
    \multicolumn{5}{@{}l}{\textbf{WMT'19}}\\
    Czech & 11 & 0.89 & 0.70 & 0.80 \\
    % German & 20 & 0.35 & -0.45 & -0.29 \\
    German & 20 & 0.14 & 0.13 & 0.16 \\
    Finnish & 12 & 0.87 & 0.82 & 0.85 \\
    Russian & 11 & 0.70 & 0.45 & 0.57 \\    
    \bottomrule
    \end{tabular}
    \caption{With a few exceptions, our grammar-based metrics correlate well with human evaluations of WMT18 and WMT19 systems. (Pearson's $r$ against Z-scores). $\dagger$:we remove outlier systems following \citet{mathur-etal-2020-tangled}. Results using \textbf{robust} Stanza parsers.}
    \label{tab:wmt_correlations_full}
    \vspace{-1em}
\end{table}

% \begin{table}[t]
%     \centering
%     \newcommand\dottedcircle{\raisebox{-.1em}{\tikz \draw [line cap=round, line width=0.2ex, dash pattern=on 0pt off 0.5ex] (0,0) circle [radius=0.79ex];}}
%     \begin{tabular}{@{}r@{}c|c|c|c@{}}
%     \toprule
%     \small{English}$\rightarrow$\dottedcircle{} \ & \#sys$^\dagger$ & $r_\textit{agree}$ & $r_\textit{as}$ & $r_\textit{agree} \cup r_\textit{as}$ \\
%     \midrule
%     \multicolumn{5}{@{}l}{\textbf{WMT'18}}\\
%     Czech & 5 & 0.90 & 0.79 & 0.86 \\
%     German & 12 & 0.15 & -0.48 & -0.43\\
%     Estonian & 12 & 0.83 & 0.59 & 0.72 \\
%     Finnish & 12 & 0.96 & 0.80 & 0.93 \\
%     Russian & 8 & 0.71 & 0.86 & 0.78 \\
%     Turkish & 7 & 0.64 & -0.30 & 0.48 \\
%     \midrule
%     \multicolumn{5}{@{}l}{\textbf{WMT'19}}\\
%     Czech & 11 & 0.89 & 0.77 & 0.87 \\
%     German & 20 & 0.34 & -0.02 & 0.04 \\
%     Finnish & 12 & 0.87 & 0.84 & 0.88 \\
%     Russian & 11 & 0.70 & 0.56 & 0.66 \\    
%     \bottomrule
%     \end{tabular}
%     \caption{With a few exceptions, our grammar-based metrics correlate well with human evaluations of WMT18 and WMT19 systems. (Pearson's $r$ against Z-scores). $\dagger$:we remove outlier systems following \citet{mathur-etal-2020-tangled}. Results using robust Stanza parsers.}
%     \label{tab:wmt_correlations}
% \end{table}

\paragraph{Correlation with Human z-scores:}

\begin{figure*}
    \begin{subfigure}{0.45\textwidth}
    \includegraphics[width=\linewidth]{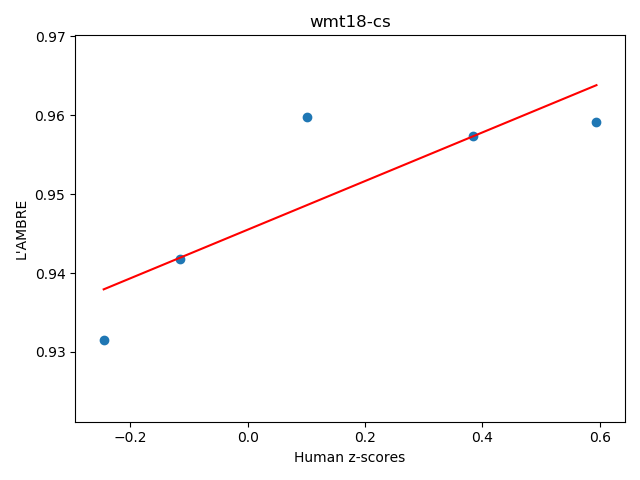}
    \caption{Czech WMT'18}
    \label{fig:wmt_human_lambre_cs_wmt18}
    \end{subfigure}
    \hfill
    \begin{subfigure}{0.45\textwidth}
    \includegraphics[width=\linewidth]{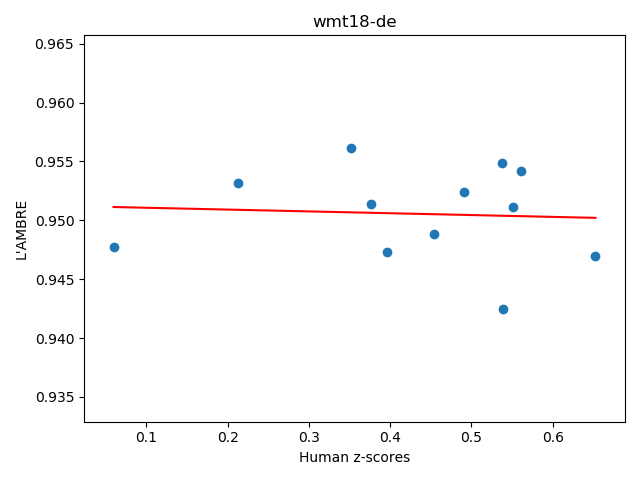}
    \caption{German WMT'18}
    \label{fig:wmt_human_lambre_de_wmt18}
    \end{subfigure}
    \begin{subfigure}{0.45\textwidth}
    \includegraphics[width=\linewidth]{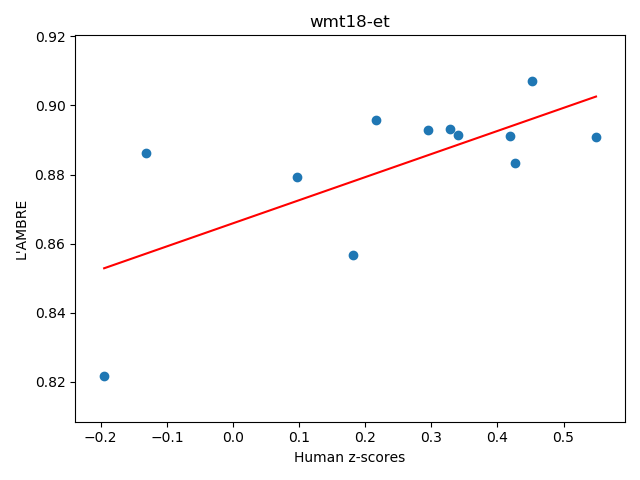}
    \caption{Estonian WMT'18}
    \label{fig:wmt_human_lambre_et_wmt18}
    \end{subfigure}
    \hfill
    \begin{subfigure}{0.45\textwidth}
    \includegraphics[width=\linewidth]{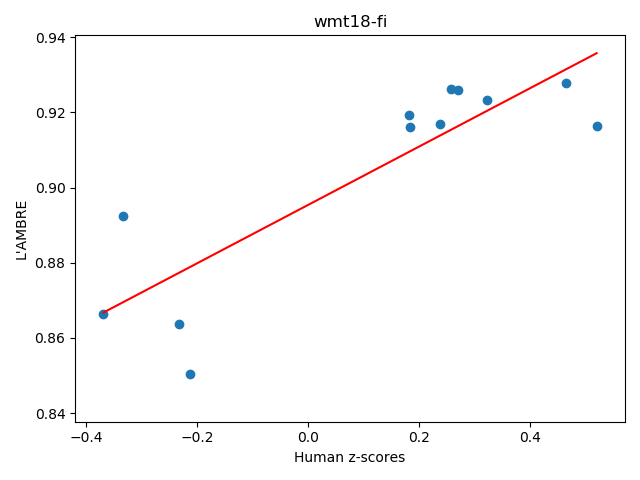}
    \caption{Finnish WMT'18}
    \label{fig:wmt_human_lambre_fi_wmt18}
    \end{subfigure}
    \begin{subfigure}{0.45\textwidth}
    \includegraphics[width=\linewidth]{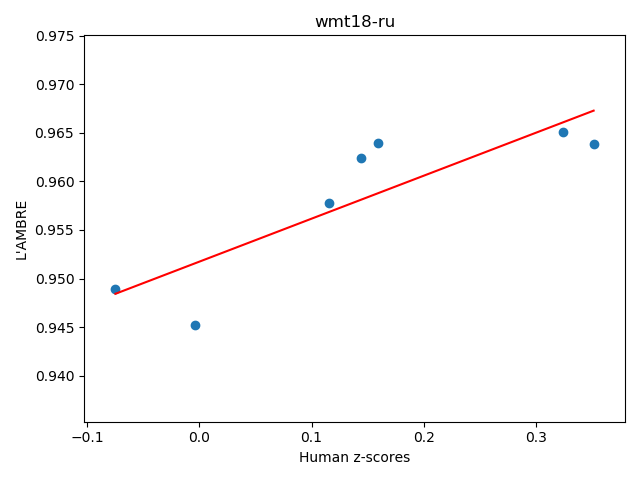}
    \caption{Russian WMT'18}
    \label{fig:wmt_human_lambre_ru_wmt18}
    \end{subfigure}
    \hfill
    \begin{subfigure}{0.45\textwidth}
    \includegraphics[width=\linewidth]{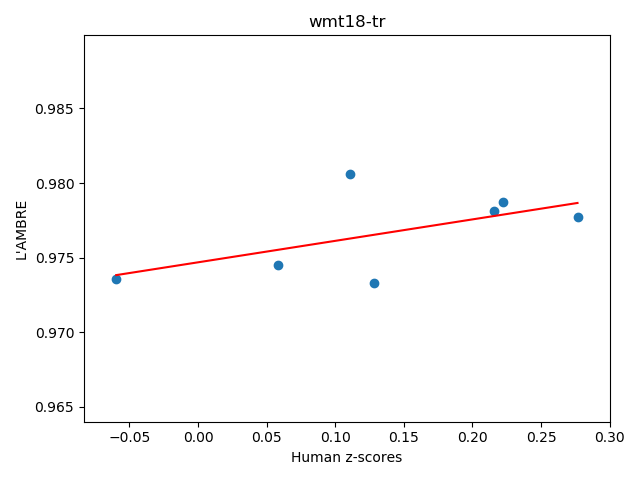}
    \caption{Turkish WMT'18}
    \label{fig:wmt_human_lambre_tr_wmt18}
    \end{subfigure}
    \caption{Scatter plot of human z-scores and \mname{} for WMT'18 systems.}
    \label{fig:wmt18_human_lambre_scatter}
\end{figure*}

\begin{figure*}
    \begin{subfigure}{0.45\textwidth}
    \includegraphics[width=\linewidth]{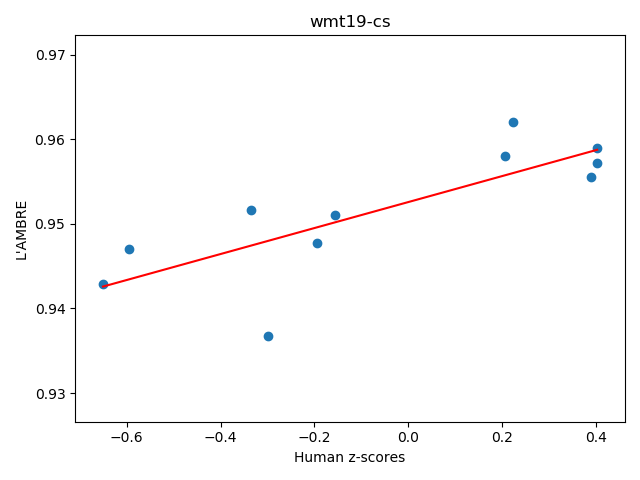}
    \caption{Czech WMT'19}
    \label{fig:wmt_human_lambre_cs_wmt19}
    \end{subfigure}
    \hfill
    \begin{subfigure}{0.45\textwidth}
    \includegraphics[width=\linewidth]{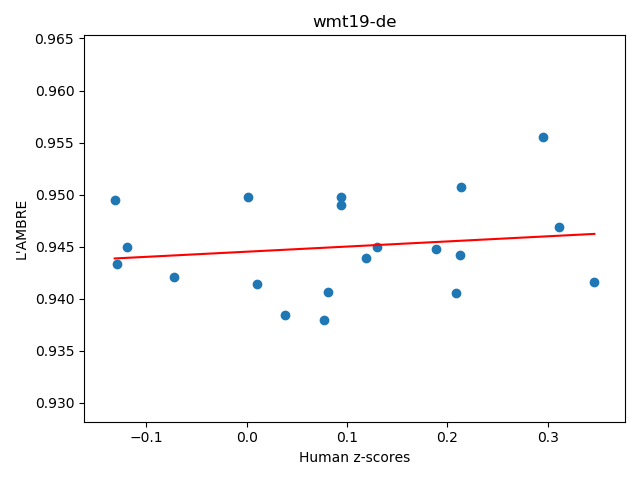}
    \caption{German WMT'19}
    \label{fig:wmt_human_lambre_de_wmt19}
    \end{subfigure}
    \begin{subfigure}{0.45\textwidth}
    \includegraphics[width=\linewidth]{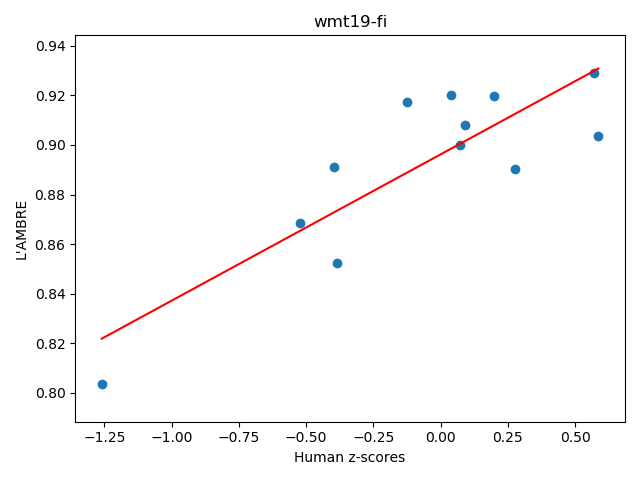}
    \caption{Finnish WMT'19}
    \label{fig:wmt_human_lambre_fi_wmt19}
    \end{subfigure}
    \hfill
    \begin{subfigure}{0.45\textwidth}
    \includegraphics[width=\linewidth]{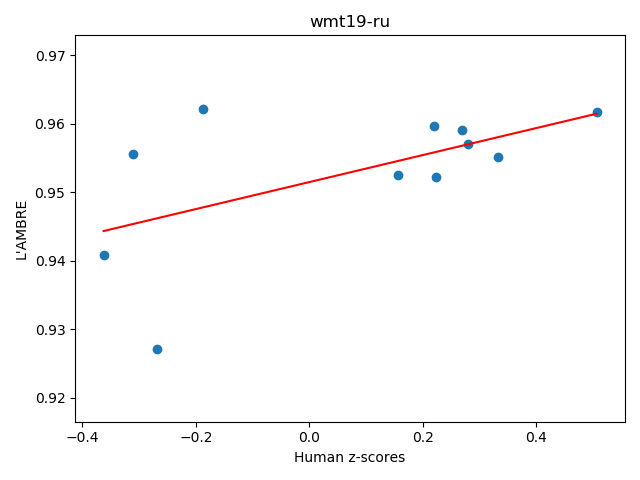}
    \caption{Russian WMT'19}
    \label{fig:wmt_human_lambre_ru_wmt19}
    \end{subfigure}
    \caption{Scatter plot of human z-scores and \mname{} for WMT'19 systems.}
    \label{fig:wmt19_human_lambre_scatter}
\end{figure*}

In \autoref{tab:wmt_correlations_full} we present a detailed account of the Pearson's r correlations between human z-scores and \mname{} for systems in WMT'18 and WMT'19. We also present the correlations with original Stanza parsers in \autoref{tab:wmt_correlations_full_original}. In \autoref{fig:wmt18_human_lambre_scatter} and \autoref{fig:wmt19_human_lambre_scatter}, we present the scatter plots comparing human z-scores and \mname{} for WMT'18 and WMT'19 respectively.

% wmt correlations using original stanza parsers

\begin{table}
    \centering
    \newcommand\dottedcircle{\raisebox{-.1em}{\tikz \draw [line cap=round, line width=0.2ex, dash pattern=on 0pt off 0.5ex] (0,0) circle [radius=0.79ex];}}
    \begin{tabular}{@{}r@{}c|c|c|c@{}}
    \toprule
    \small{English}$\rightarrow$\dottedcircle{} \ & \#sys$^\dagger$ & $r_\textit{agree}$ & $r_\textit{as}$ & $r_\textit{agree} \cup r_\textit{as}$ \\
    \midrule
    \multicolumn{5}{@{}l}{\textbf{WMT'18}}\\
    Czech & 5 & 0.88 & 0.85 & 0.87 \\
    German & 12 & -0.08 & 0.02 & 0.04 \\
    Estonian & 12 & 0.84 & 0.68 & 0.74 \\
    Finnish & 12 & 0.96 & 0.73 & 0.85 \\
    Russian & 7 & 0.66 & 0.84 & 0.90 \\
    Turkish & 7 & 0.64 & -0.86 & 0.51 \\
    \midrule
    \multicolumn{5}{@{}l}{\textbf{WMT'19}}\\
    Czech & 11 & 0.86 & 0.46 & 0.63 \\
    German & 20 & 0.44 & 0.10 & 0.17 \\
    Finnish & 12 & 0.85 & 0.81 & 0.84 \\
    Russian & 11 & 0.74 & 0.58 & 0.69 \\    
    \bottomrule
    \end{tabular}
    \caption{Correlations with human evaluations of WMT18 and WMT19 systems. (Pearson's $r$ against Z-scores). $\dagger$:we remove outlier systems following \citet{mathur-etal-2020-tangled}. Results using \textbf{original} Stanza parsers.}
    \label{tab:wmt_correlations_full_original}
    % \vspace{-1em}
\end{table}

\begin{table}[ht]
    \centering
    \newcommand\dottedcircle{\raisebox{-.1em}{\tikz \draw [line cap=round, line width=0.2ex, dash pattern=on 0pt off 0.5ex] (0,0) circle [radius=0.79ex];}}
    \begin{tabular}{@{}r@{}c|c|c@{}}
    \toprule
    \small{English}$\rightarrow$\dottedcircle{} \ & \#sys$^\dagger$ & BLEU & chrF \\
    \midrule
    \multicolumn{4}{@{}l}{\textbf{WMT'18}}\\
    Czech & 5 & 0.83 & 0.81 \\
    German & 12 & 0.18 & 0.14 \\
    Estonian & 12 & 0.75 & 0.66 \\
    Finnish & 12 & 0.89 & 0.86 \\
    Russian & 7 & 0.85 & 0.84 \\
    Turkish & 7 & 0.61 & 0.66 \\
    \midrule
    \multicolumn{4}{@{}l}{\textbf{WMT'19}}\\
    Czech & 11 & 0.85 & 0.83 \\
    German & 20 & 0.01 & 0.01 \\
    Finnish & 12 & 0.86 & 0.85 \\
    Russian & 11 & 0.65 & 0.53 \\    
    \bottomrule
    \end{tabular}
    \caption{Correlations with BLEU, chrF for WMT'18 and WMT'19 systems using all the rules. (Pearson's $r$ against \mname{}). $\dagger$:we remove outlier systems following \citet{mathur-etal-2020-tangled}. Results using \textbf{robust} Stanza parsers.}
    \label{tab:wmt_correlations_bleu}
\end{table}

\paragraph{Correlation with other metrics}

\begin{figure*}
    \begin{subfigure}{0.45\textwidth}
    \includegraphics[width=\linewidth]{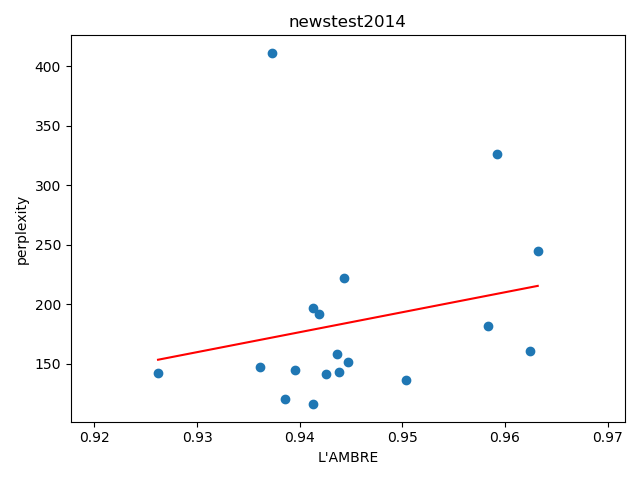}
    \caption{German WMT'14}
    \label{fig:wmt_ppl_lambre_de_wmt14}
    \end{subfigure}
    \hfill
    \begin{subfigure}{0.45\textwidth}
    \includegraphics[width=\linewidth]{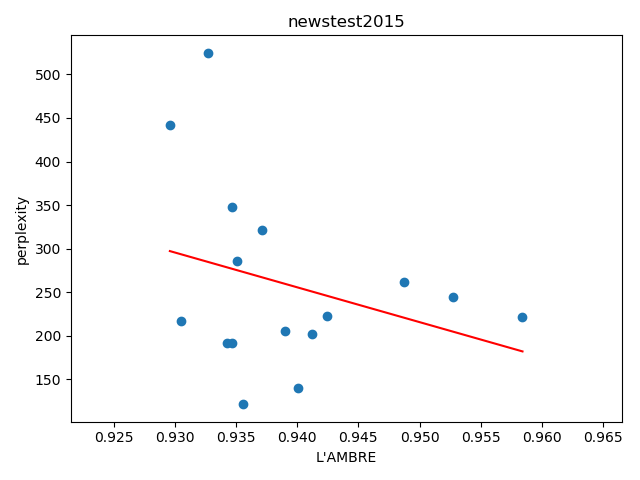}
    \caption{German WMT'15}
    \label{fig:wmt_ppl_lambre_de_wmt15}
    \end{subfigure}
    \begin{subfigure}{0.45\textwidth}
    \includegraphics[width=\linewidth]{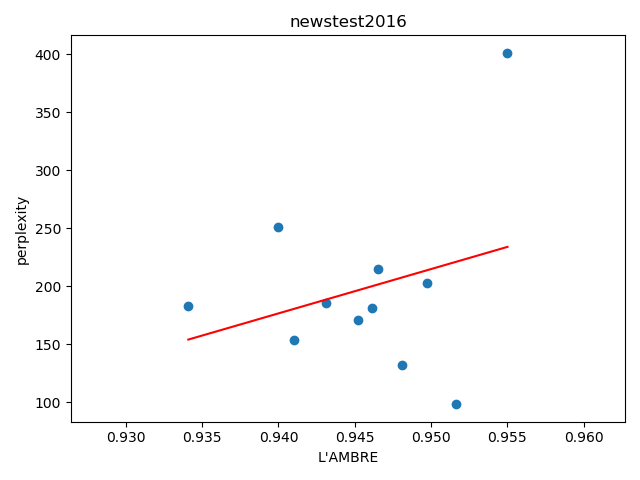}
    \caption{German WMT'16}
    \label{fig:wmt_ppl_lambre_de_wmt16}
    \end{subfigure}
    \hfill
    \begin{subfigure}{0.45\textwidth}
    \includegraphics[width=\linewidth]{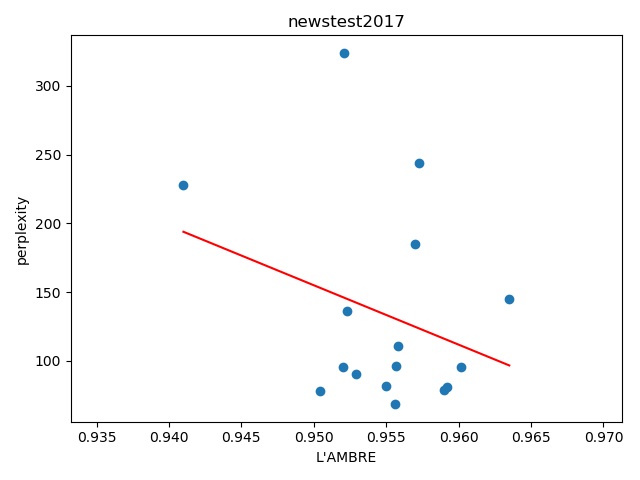}
    \caption{German WMT'17}
    \label{fig:wmt_ppl_lambre_de_wmt17}
    \end{subfigure}
    \begin{subfigure}{0.45\textwidth}
    \includegraphics[width=\linewidth]{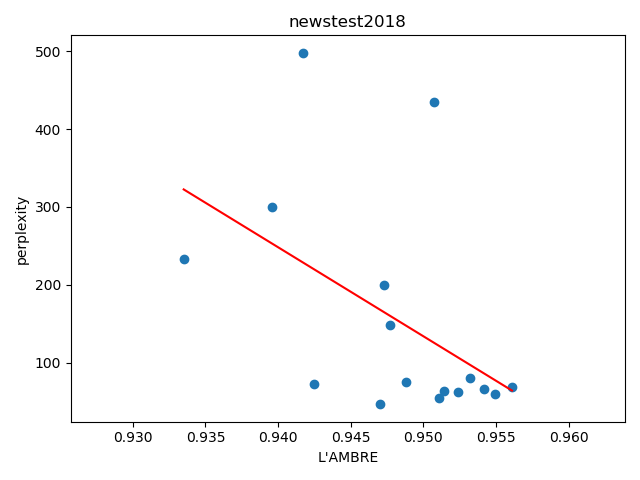}
    \caption{German WMT'18}
    \label{fig:wmt_ppl_lambre_de_wmt18}
    \end{subfigure}
    \hfill
    \begin{subfigure}{0.45\textwidth}
    \includegraphics[width=\linewidth]{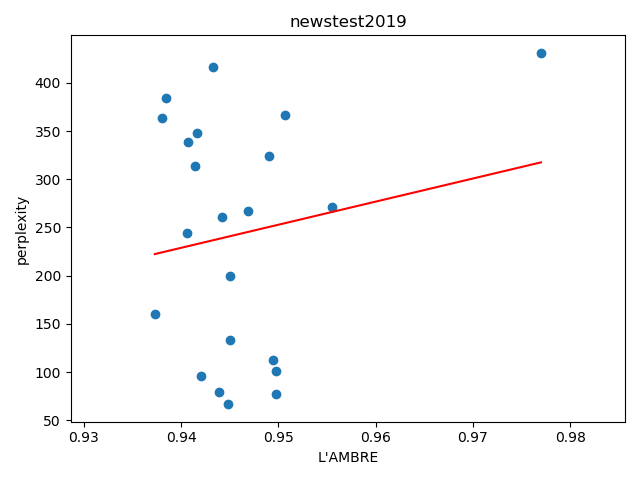}
    \caption{German WMT'19}
    \label{fig:wmt_ppl_lambre_de_wmt19}
    \end{subfigure}
    \caption{Scatter plot of perplexity and \mname{} for German WMT systems from WMT'14$\rightarrow$'19. High \mname{} and low perplexity indicate better systems. As expected, we see a negative correlation between the two metrics for WMT'15, WMT'17, and WMT'18. But for WMT'14, WMT'16, and WMT'19, we see positive correlations, indicating potential limitations of our metric.}
    \label{fig:wmt_ppl_lambre_de}
\end{figure*}

\begin{figure*}
    \begin{subfigure}{0.45\textwidth}
    \includegraphics[width=\linewidth]{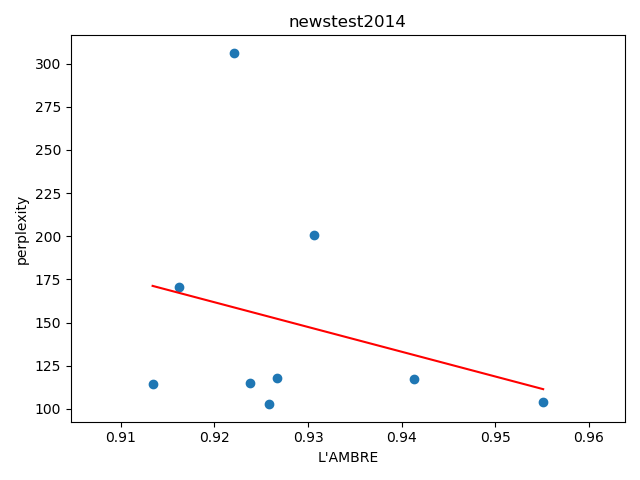}
    \caption{Russian WMT'14}
    \label{fig:wmt_ppl_lambre_ru_wmt14}
    \end{subfigure}
    \hfill
    \begin{subfigure}{0.45\textwidth}
    \includegraphics[width=\linewidth]{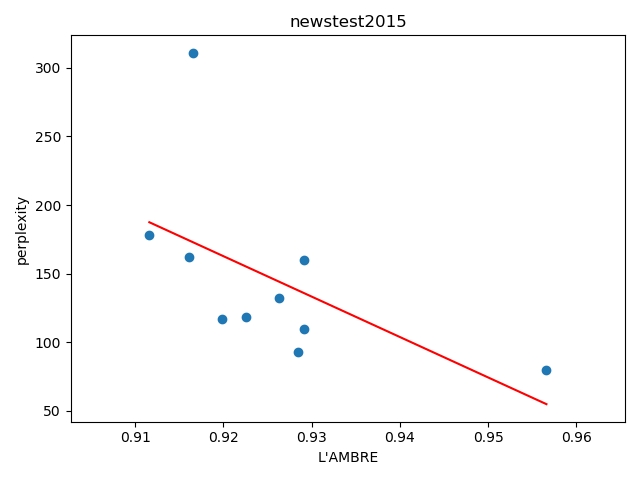}
    \caption{Russian WMT'15}
    \label{fig:wmt_ppl_lambre_ru_wmt15}
    \end{subfigure}
    \begin{subfigure}{0.45\textwidth}
    \includegraphics[width=\linewidth]{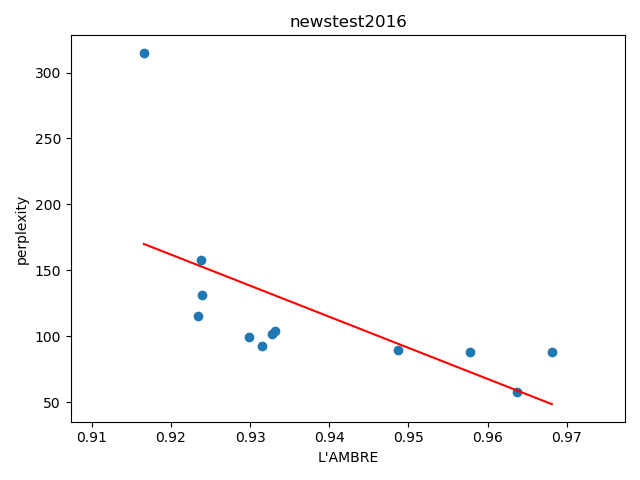}
    \caption{Russian WMT'16}
    \label{fig:wmt_ppl_lambre_ru_wmt16}
    \end{subfigure}
    \hfill
    \begin{subfigure}{0.45\textwidth}
    \includegraphics[width=\linewidth]{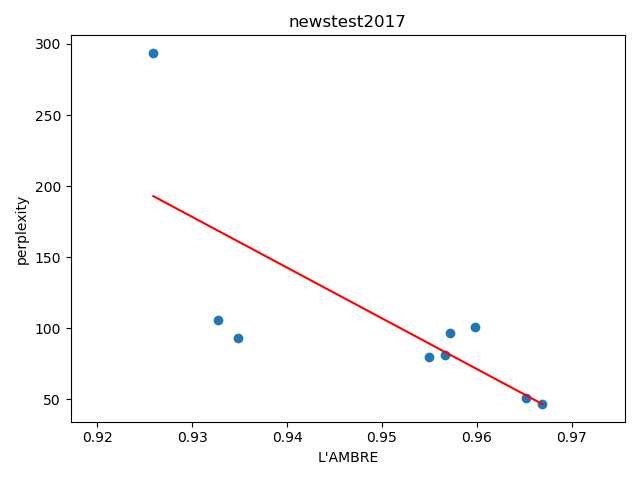}
    \caption{Russian WMT'17}
    \label{fig:wmt_ppl_lambre_ru_wmt17}
    \end{subfigure}
    \begin{subfigure}{0.45\textwidth}
    \includegraphics[width=\linewidth]{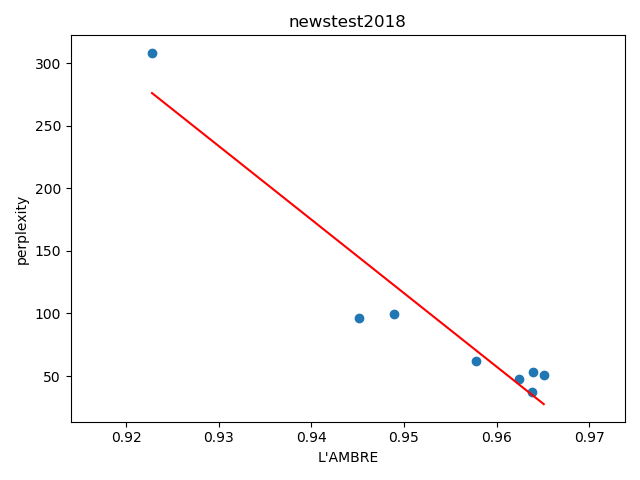}
    \caption{Russian WMT'18}
    \label{fig:wmt_ppl_lambre_ru_wmt18}
    \end{subfigure}
    \hfill
    \begin{subfigure}{0.45\textwidth}
    \includegraphics[width=\linewidth]{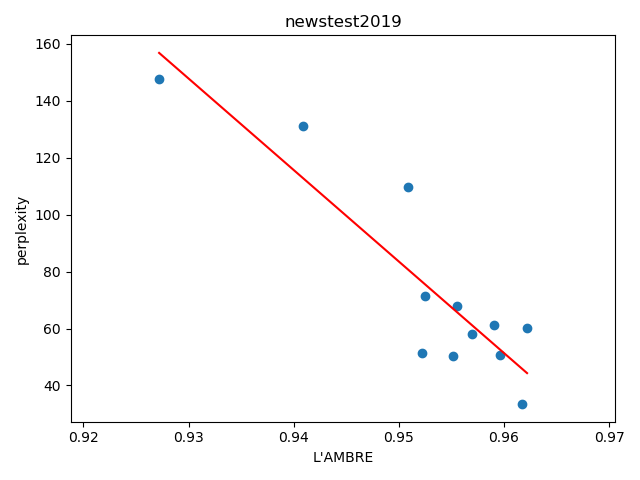}
    \caption{Russian WMT'19}
    \label{fig:wmt_ppl_lambre_ru_wmt19}
    \end{subfigure}
    \caption{Scatter plot of perplexity and \mname{} for Russian WMT systems from WMT'14-'19. High \mname{} and low perplexity indicate better systems. As expected, we see negative correlation between the two metrics across the years.}
    \label{fig:wmt_ppl_lambre_ru}
\end{figure*}

In \autoref{tab:wmt_correlations_bleu} we present a comparison of \mname{} with BLEU~\cite{papineni-etal-2002-bleu} and chrF~\cite{popovic-2015-chrf}. In \autoref{fig:wmt_ppl_lambre_de} and \autoref{fig:wmt_ppl_lambre_ru}, we present scatter plots comparing perplexity and \mname{} for WMT systems from WMT'14 to WMT'19. To use perplexity as a corpus fluency measure, we first compute perplexity of each output translation and then take an average over all sentences in the target test set to obtain a corpus perplexity score for each WMT system. On most occasions, as expected, we see a negative correlation between perplexity and \mname{}, more strongly in Russian than in German.

\subsection{Diachronic analysis of WMT systems}
\label{ssec:appendix_diachronic_wmt}

\autoref{fig:diachronic_results_appendix} presents a diachronic study of WMT systems for Czech, Finnish, and Turkish using \mname{}. \autoref{fig:diachronic_results_rules_appendix} shows the morpho-syntactic rule specific trends for Russian WMT.

\subsection{Rule Extraction Statistics}

For extracting agreement ($r_\textit{agree}$), case assignment and verb form choice ($r_\textit{as}$) rules, we use the largest available treebank for the language from SUD. \autoref{tab:rule_extraction_count_statistics} presents the rule counts for the languages discussed in this paper.

\begin{table}[ht]
\centering
\begin{tabular}{@{}lcc@{}}
\toprule
Treebank & \# $r_\textit{agree}$ & \# $r_\textit{as}$ \\
\midrule
Czech-PDT & 28 & 33 \\
German-HDT & 30 & 23 \\
Greek-GDT & 11 & 12 \\
Estonian-EDT & 22 & 31 \\
Finnish-TDT & 19 & 35 \\
Russian-SynTagRus & 25 & 35 \\
Turkish-IMST & 32 & 6 \\
\bottomrule
\end{tabular}
\caption{Statistics of rules extracted from SUD treebanks.}
\label{tab:rule_extraction_count_statistics}
\end{table}

\subsection{Reproducibility Checklist}

\subsubsection{Model Training}

For training robust dependency parsers, we use the training infrastructure provided by Stanza authors.\footnote{\url{https://stanfordnlp.github.io/stanza/training.html}} We use the same set of language-specific hyperparameters as the original Stanza parsers and taggers. All our training is performed on a single GeForce RTX 2080 GPU.

\subsubsection{Resources}

In this work we use WMT metrics dataset,\footnote{\url{http://www.statmt.org/wmt19/metrics-task.html}} WMT human evaluation scores,\footnote{\url{http://www.statmt.org/wmt19/results.html}} SUD treebanks,\footnote{\url{https://surfacesyntacticud.github.io/data/}} UD2SUD converter.\footnote{\url{https://github.com/surfacesyntacticud/tools}}

\pgfplotstableread[row sep=\\,col sep=&]{
wmt & agr-conj-NOUN-NOUN:Case & agr-det-DET-NOUN:Gender & agr-flat@name-PROPN-PROPN:Number & agr-mod-ADJ-NOUN:Case & agr-mod-VERB-NOUN:Case & agr-mod-VERB-NOUN:Gender & agr-mod-VERB-NOUN:Number & agr-subj-NOUN-VERB:Gender & agr-subj-NOUN-VERB:Number & agr-subj-PRON-VERB:Person & agr-subj@pass-NOUN-VERB:Number & args-comp:obj-NOUN-VERB:depd:Case & args-comp:obj@x-VERB-ADJ:depd:VerbForm & args-comp:obj@x-VERB-VERB:depd:VerbForm & args-subj-NOUN-AUX:depd:Case & args-subj-NOUN-VERB:depd:Case & args-subj@pass-NOUN-AUX:depd:Case & args-udep-PRON-VERB:depd:Case & args-udep-VERB-NOUN:depd:VerbForm\\ 
'14 & 0.7789 & 0.9894 & 0.9974 & 0.9258 & 0.8244 & 0.9050 & 0.8525 & 0.9202 & 0.8719 & 0.9930 & 0.8592 & 0.9522 & 1.0000 & 1.0000 & 0.9521 & 0.9389 & 0.8909 & 0.8684 & 1.0000 \\ 
'15 & 0.7760 & 0.9861 & 1.0000 & 0.9123 & 0.8364 & 0.9041 & 0.8628 & 0.9085 & 0.8584 & 0.9884 & 0.8110 & 0.9522 & 1.0000 & 1.0000 & 0.9201 & 0.9275 & 0.9006 & 0.9028 & 1.0000 \\ 
'16 & 0.8088 & 0.9906 & 1.0000 & 0.9405 & 0.8392 & 0.9199 & 0.8779 & 0.9252 & 0.8700 & 0.9902 & 0.8521 & 0.9684 & 1.0000 & 1.0000 & 0.9451 & 0.9232 & 0.8889 & 0.9221 & 1.0000 \\ 
'17 & 0.8848 & 0.9900 & 1.0000 & 0.9714 & 0.9345 & 0.9300 & 0.9500 & 0.9670 & 0.9275 & 0.9972 & 0.9241 & 0.9804 & 1.0000 & 1.0000 & 0.9556 & 0.9364 & 0.8900 & 0.9259 & 1.0000 \\ 
'18 & 0.8962 & 0.9930 & 1.0000 & 0.9745 & 0.9332 & 0.9336 & 0.9470 & 0.9800 & 0.9378 & 0.9955 & 0.9367 & 0.9840 & 1.0000 & 1.0000 & 0.9619 & 0.9398 & 0.8846 & 0.9272 & 1.0000 \\ 
'19 & 0.8636 & 0.9928 & 1.0000 & 0.9758 & 0.9268 & 0.9424 & 0.9423 & 0.9718 & 0.9362 & 0.9972 & 0.9315 & 0.9873 & 1.0000 & 1.0000 & 0.9749 & 0.9386 & 0.9032 & 0.8734 & 1.0000 \\ 
}\diachronicdata

\input{images/wmt}

\begin{figure*}[ht]
\centering
\begin{subfigure}{0.48\textwidth}
\resizebox{\textwidth}{!}{
\begin{tabular}{p{3.8cm}p{3.8cm}}
\wmtplot{\csdata}{\csavg}{Czech}{.85} & 
\wmtplot{\fidata}{\fiavg}{Finnish}{.75}\\
\wmtplot{\trdata}{\travg}{Turkish}{.93} & \\
\multicolumn{2}{c}{\small $\blacksquare$: systems average \ \ \textcolor{red}{$\circ$}: system \ \ $\bullet$: reference}\\
\end{tabular}}
\caption{A diachronic study of grammatical well-formedness of WMT English$\rightarrow$X systems' outputs. The systems in general are becoming more fluent with very passing year. In the last two years the best systems produce as well-formed outputs as the reference translations.} %\ap{should we report median or mean score?}}
\label{fig:diachronic_results_appendix}
\end{subfigure}
\hfill
\begin{subfigure}{0.48\textwidth}
\resizebox{\textwidth}{!}{
\begin{tabular}{m{3.6cm}m{3.6cm}}
\begin{tabular}{c}
\tikplotmiddle{agr-mod-VERB-NOUN:Number}{(a) {$r_{agree}(\mathtt{Verb,Noun,mod})\!\rightarrow\mathtt{Number}$}}\\
\tikplotmiddle{agr-mod-VERB-NOUN:Gender}{(d) $r_{agree}(\mathtt{Verb,Noun,mod})\!\rightarrow\!\mathtt{Gender}$}\\
\tikplotmiddle{args-subj-NOUN-AUX:depd:Case}{(h) $r_{as}(\mathtt{N,Aux,subj})\!\rightarrow\!\mathtt{N}_{\mathtt{Case}}\mathtt{=Nom}$}\\
\end{tabular}
&
\begin{tabular}{c}
\tikplot{agr-mod-VERB-NOUN:Case}{(c) $r_{agree}(\mathtt{Verb,Noun,mod})\!\rightarrow\!\mathtt{Case}$}
\\
\tikplotmiddle{agr-mod-ADJ-NOUN:Case}{(e) $r_{agree}(\mathtt{Adj,Noun,mod})\!\rightarrow\!\mathtt{Case}$}\\
\tikplotmiddle{args-udep-PRON-VERB:depd:Case}{(j) $r_{as}(\mathtt{Pron,V,udep})\!\!\rightarrow\!\!\mathtt{Pron}_{\mathtt{Case}}\!=\!\mathtt{Gen,\!Ins}$}\\
\end{tabular}
\end{tabular}}
% \vspace{-2em}
\caption{Diachronic analysis of additional agreement ($r_{agree}$) and argument structure ($r_{as}$) rules in Russian WMT. We report the \textit{median} well-formedness score for each WMT year.}
\label{fig:diachronic_results_rules_appendix}
\end{subfigure}
\caption{Diachronic study of the grammatical well-formedness of WMT systems.}
\end{figure*}


\end{document}